%% file: main.tex
\documentclass[sigconf]{acmart}

\usepackage{array}
\usepackage{booktabs, graphicx, makecell}
\usepackage{CJKutf8}
\usepackage{cuted}

\AtBeginDocument{%
  }

\setcopyright{acmlicensed}
\copyrightyear{2018}
\acmYear{2018}
\acmDOI{XXXXXXX.XXXXXXX}

\acmConference[Conference acronym 'XX]{Make sure to enter the correct
  conference title from your rights confirmation email}{June 03--05,
  2018}{Woodstock, NY}
\acmISBN{978-1-4503-XXXX-X/2018/06}

\makeatletter
\newcommand{\authornotetext}[2]{%
  \g@addto@macro\@authornotes{\footnotetext[#1]{#2}}}
\makeatother
\begin{document}

\title[Learning to Allocate Incentives for Incentivized Advertising]{Learning to Allocate Incentives for Incentivized Advertising via Offline Model-Based Reinforcement Learning}

\author{Zilin Zhao}
\authornotemark[2]
\affiliation{%
  \institution{State Key Laboratory of Novel Software Technology, Nanjing University}
  \city{Nanjing}
  \country{China}}
\affiliation{%
  \institution{ByteDance}
  \city{Beijing}
  \country{China}}
\email{zhao\_zilin@outlook.com}

\author{Han Yang}
\authornotemark[2]
\affiliation{%
  \institution{State Key Laboratory of Novel Software Technology, Nanjing University}
  \city{Nanjing}
  \country{China}}
\affiliation{%
  \institution{ByteDance}
  \city{Beijing}
  \country{China}}
\email{yh17860742238@163.com}

\author{Tianpei Yang}
\authornotemark[1]
\affiliation{%
  \institution{State Key Laboratory of Novel Software Technology, Nanjing University}
  \city{Nanjing}
  \country{China}}
\email{tianpei.yang@nju.edu.cn}

\author{Fangsheng Huang}
\authornotemark[1]
\affiliation{%
  \institution{ByteDance}
  \city{Beijing}
  \country{China}}
\email{hfs91@sina.com}

\author{Yanfei Cui}
\affiliation{%
  \institution{ByteDance}
  \city{Beijing}
  \country{China}}
\email{cuiyanfei.97@bytedance.com}

\author{Kan Peng}
\affiliation{%
  \institution{ByteDance}
  \city{Beijing}
  \country{China}}
\email{pengkan@bytedance.com}

\author{Yi Li}
\affiliation{%
  \institution{ByteDance}
  \city{Beijing}
  \country{China}}
\email{liyi.lynne@bytedance.com}

\author{Yiming Zong}
\affiliation{%
  \institution{The Hong Kong University of Science and Technology}
  \city{Hong Kong}
  \country{China}}
\affiliation{%
  \institution{ByteDance}
  \city{Beijing}
  \country{China}}
\email{yzongac@connect.ust.hk}

\author{Hao Zhang}
\author{Yinsong Xue}
\affiliation{%
  \institution{ByteDance}
  \city{Beijing}
  \country{China}}
\email{zhang.hao.alan@gmail.com}
\email{xueyinsong@bytedance.com}

\authornotetext{2}{These authors contributed equally to this work.}
\authornotetext{1}{Corresponding authors.}

\renewcommand{\shortauthors}{Zhao et al.}

\begin{abstract}
Complete your ad view and grab a 5-cent bonus! In incentivized advertising, a platform offers users a bonus before it observes the downstream ad revenue. The promise encourages users to click and watch ads to completion. The platform must balance the incentive cost promised in advance with the ad revenue observed afterward. Low incentives may lead users to reject the offer, causing the platform to lose monetization opportunities. High incentives may erode realized net profit. Beyond budget constraints, the platform also needs to consider how current incentives shape user expectations and subsequent engagement. Incentive allocation is therefore a sequential decision-making problem with revenue observed afterward, cost sensitivity, and carryover effects.

Current literature has not yet studied decision-making algorithms for this setting. Related work, such as auto-bidding and targeted promotion, covers only part of the problem: it either assumes existing ad opportunities or optimizes user incentives outside the ad monetization pipeline. To model user feedback and learn a cost-controllable sequential incentive policy, we formulate this problem as an MDP and adopt an offline model-based RL algorithm to solve it. We first learn a world model for user feedback and ad revenue, and then combine it with conservative policy optimization. An independent counterfactual evaluation scorer tests each newly learned incentive-allocation policy on held-out logs, helping identify policies that are expected to improve net profit before exposing costly live traffic. We validate the framework on large-scale industrial data and online A/B tests. The scorer provides a stable offline signal for pre-launch policy selection. Online A/B tests trace a deployment path from causal inference to offline RL and then to Offline-MBRL: our MB IQL method improves per-user net profit by 7.96\% over TD3+BC, while reversing to plain IQL reduces net profit by 6.56\%. Both have p<0.0001.
\end{abstract}



\maketitle

\begin{strip}
  \centering
  \includegraphics[width=\textwidth]{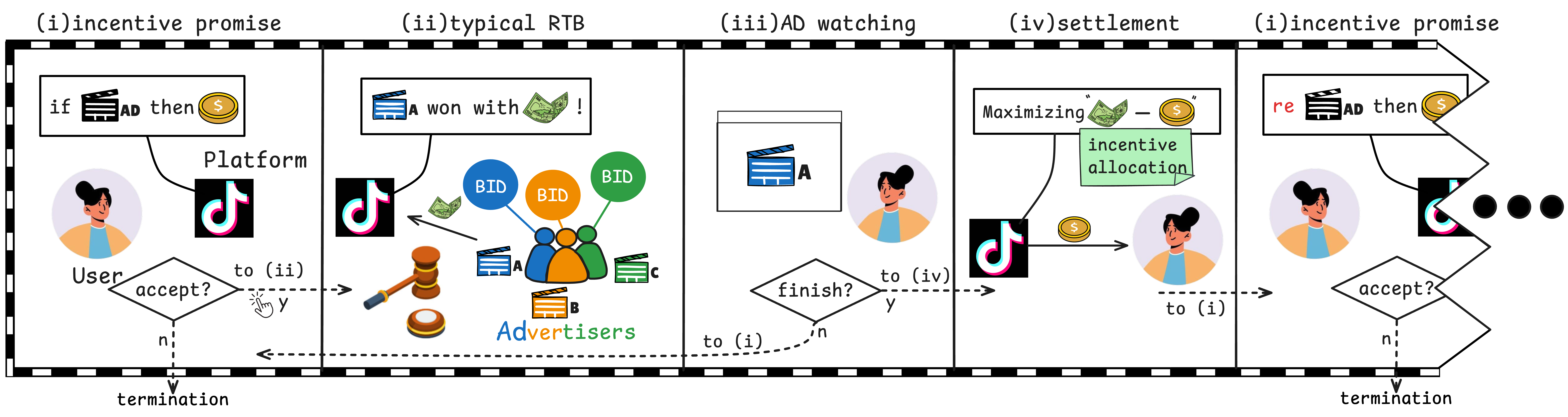}
  \captionof{figure}{Incentivized advertising commits user incentives before downstream ad revenue is observed. (i) The platform first makes an incentive promise to the app user, who may accept or reject the offer. (ii) An accepted offer initiates the corresponding real-time bidding (RTB) process, through which the platform realizes advertiser revenue. (iii) The winning ad is shown to the user, who may complete or abandon the watch. (iv) After completion, the promised incentive is settled, and the platform starts another loop with a new incentive promise.}

  \Description{An infographic showing interactions among the user, platform, and advertiser in incentivized advertising. The platform offers an incentive amount $a$. User acceptance triggers an exposure, the associated RTB request, and RTB billing. RTB revenue $I$ is realized when the exposure is generated, and the promised incentive is paid only after completion. User interest, fatigue, history, cooldown, expectations, and strategic response affect the next state.}
  \label{fig:pipeline}
\end{strip}



\section{Introduction}

\textbf{Incentivized advertising} is an advertising product in which a platform offers users incentives for ad watching. The platform offers a direct or indirect bonus, such as cash or virtual currency, to encourage users to click and complete an ad view. Traditional feed ads and app-opening ads mainly compete for attention through placement, relevance, and creative content. In contrast, incentivized advertising pays users for completing an ad view.

\textbf{Incentive allocation} balances the incentive cost promised in advance against the ad revenue observed afterward. When the platform shows an incentive offer, it does not know whether an exposure will be generated. Once it is generated, advertiser revenue is determined by the associated real-time bidding (RTB) process. If the incentive is too low, users may reject the offer, causing the platform to lose monetization opportunities. If the incentive is too high, users may click more to bring income, but the added cost may exceed the added revenue. A high incentive may also raise users' expectations on future bonuses. Incentive allocation must also account for cost limits and delivery targets. It is therefore a cost-sensitive sequential decision problem,  with delayed revenue and short-horizon carryover effects. To the best of our knowledge, public literature has not jointly studied this setting, where incentives create ad exposures, and downstream RTB revenue are under short-horizon sequential dynamics.

\textbf{Existing related research} is scattered across adjacent areas, but each covers only part of this problem. Auto-bidding studies advertiser-side bidding algorithms in RTB, where the problem is also sequential and budget-sensitive. Representative auto-bidding methods include RL-based~\cite{rtb_rl,drlb} and generative algorithms~\cite{aigb,gave}. However, the object being bid on is an already-existing exposure provided by the platform. These methods do not need to model whether an exposure itself will be generated. In fact, auto-bidding corresponds to the revenue side of the problem studied in this paper. Targeted promotion studies personalized coupons, subsidies, or cash incentives for users. Common targeted-promotion methods include causal inference~\cite{metalearners_cate,causal_forests,e3ir,UV}, operations research~\cite{unified}, and reinforcement learning~\cite{bcrlsp}. Targeted promotion treats incentives as treatments on users, and must trade off user response against incentive cost. Nevertheless, existing work usually uses incentives to improve conversion, retention, or lifetime value (LTV) over existing user demand, without considering the advertising monetization pipeline. In incentivized advertising, however, the incentive first determines whether an exposure is generated at all. If no exposure is generated, there is no ad revenue. Therefore, incentive allocation is neither bidding for existing exposures nor promoting conversion over existing user demand. Instead, it uses user incentives to create exposures and optimizes their allocation by accounting for within-window user-response carryover and incentive cost.

\textbf{Reinforcement learning (RL)} is naturally suited for budget constraints and sequential decision-making, and has been widely applied to auto-bidding~\cite{rtb_rl,drlb}, targeted promotion~\cite{bcrlsp}, and recommendation~\cite{deep_page,topk_offpolicy,onetrans}. However, directly deploying RL in incentivized advertising faces challenges. Online RL is impractical because each exploration incurs real incentive spending, and a policy that has not yet converged may simultaneously cause platform losses and user-experience fluctuations. Offline RL avoids risky online exploration, but logged data are constrained by historical delivery policies, traffic allocation, and cost-control mechanisms. As a result, incentive amounts in the logs are often sparsely covered. This can lead to out-of-distribution (OOD) action overestimation and difficult counterfactual evaluation~\cite{cql,iql,td3bc}. It also makes complex user behavior hard to capture through value-function learning alone.

To address these challenges, we propose an Offline Model-Based Reinforcement Learning (Offline-MBRL) framework for incentive allocation in incentivized advertising. The RL formulation captures the bounded-window sequence of incentive attempts and allows cost-sensitive learning.
The model-based component structures sparse logged feedback into response gates, posterior ad revenue, and local state transitions, making the incentivized-advertising environment learnable around logged states.
Conservative policy optimization limits the policy's ability to exploit model errors. Because industrial policy development repeatedly compares cost regimes and training checkpoints, we attach an independent Counterfactual Evaluation Scorer (CES) to each logged dataset for reusable, per-user KPI screening before costly online deployment.
We deploy the learned policy in a real incentivized advertising system and evaluate it through CES-based offline experiments and online A/B tests. The main contributions are as follows:

\begin{enumerate}
  \item We identify and formulate incentive allocation in incentivized advertising as a new cost-sensitive sequential decision problem, where user incentives create monetizable ad opportunities before the platform observes ad revenue.
  \item We propose an Offline-MBRL framework for this problem. The framework learns a user-response world model from offline logs enriched with request-level context, performs conservative model-based policy optimization, and uses an independent CES for offline policy screening. This enables the platform to optimize user-side incentives without high-risk online RL exploration.
  \item We validate the proposed method on large-scale industrial data, ablation studies, and online A/B tests. Results show that the method improves platform net profit while controlling incentive cost, and that the offline CES ranking is consistent with the sequential online deployment evidence. We also verify that the extensions remain viable on a public benchmark.
\end{enumerate}

\section{Incentivized Advertising}


\textbf{Business pipeline and downstream payoff.}
As illustrated in Fig.~\ref{fig:pipeline}, a typical incentivized advertising process starts before the downstream ad revenue is known. The platform first shows an incentive message to the user and promises a bonus with monetary value $a$ after ad completion. For consistent modeling and reporting, we use \textbf{\emph{exposure}} to denote a successful serving event: in our production logs, the user-acceptance signal, resulting ad impression, associated RTB request, and RTB billing record are aligned observations of that event. If the incentive attempt does not generate an exposure, the current round ends directly: no ad revenue is realized, and no incentive cost is paid.

If the incentive attempt generates an exposure, advertisers participate in the associated RTB process and the platform observes revenue $I$ here. However, the user receives the promised amount $a$ only after completing the ad view. Thus, the exposure determines revenue, whereas completion triggers the incentive payment.


Incentive allocation uses incentives to create potential exposures. A low incentive may fail to generate an exposure. A high incentive may increase exposure generation or completion but make net payoff negative after the payout. The platform must allocate incentives to opportunities likely to generate positive net value under this delayed-revenue structure.

\begin{figure}[!t]
  \centering
  \includegraphics[width=\linewidth]{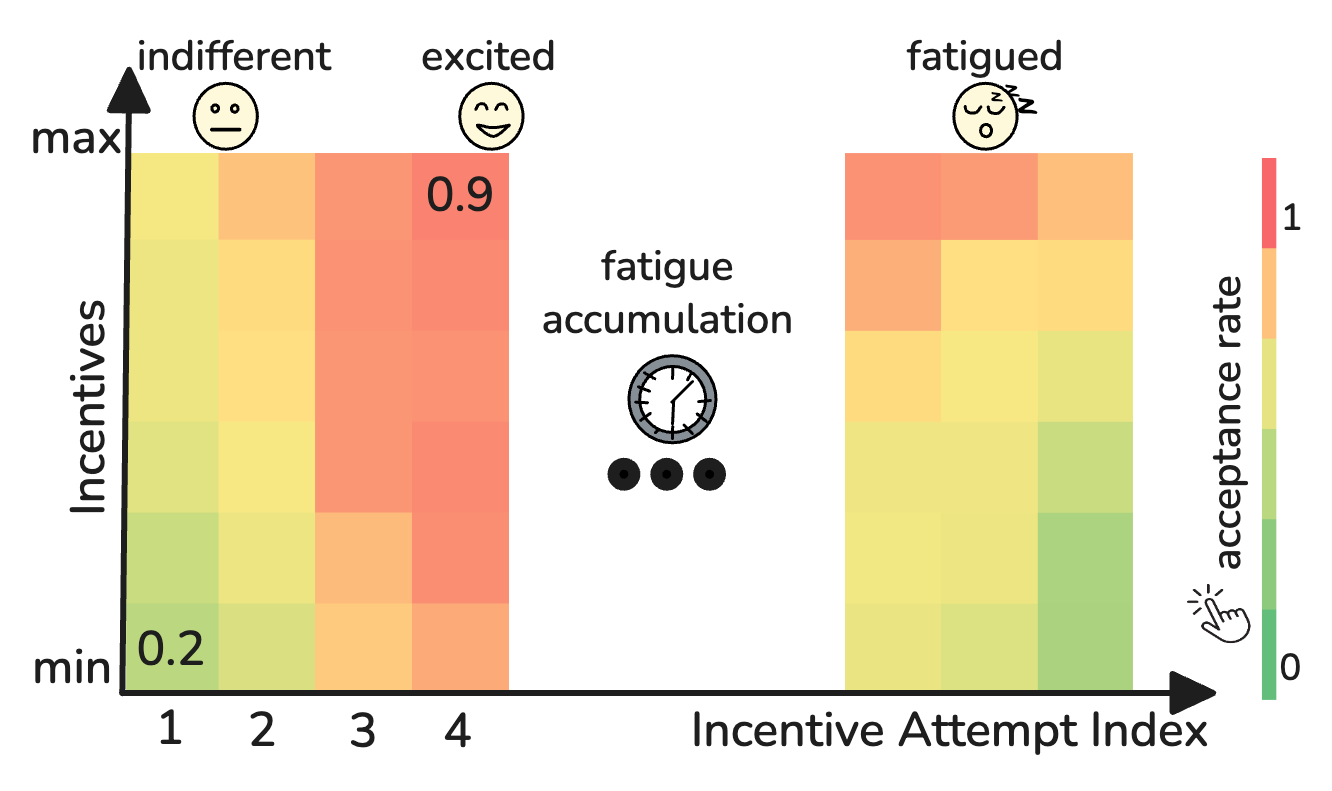}
  \caption{User feedback in incentivized advertising exhibits short-window carryover patterns. Users are often indifferent before the first incentive, so the initial acceptance rate is relatively low. Once accepted, users may enter a short excited stage with higher click propensity. As incentive fatigue accumulates, the willingness to click gradually declines. The data are aggregated from the real interaction trajectories of 100 randomly selected users.}
  \Description{Illustration of the indifferent, excited, and fatigue stages in incentivized advertising. The figure shows that the exposure rate starts low, rises after the first exposure, and then decreases as incentive fatigue accumulates. }
  \label{fig:user_seqential}
\end{figure}

\textbf{Bounded-window user dynamics.}
The difficulty of incentive allocation is further amplified by user behavior. Although the immediate outcome can be summarized by whether the exposure is generated, it is shaped by user interest, incentive fatigue, historical incentive experience, and expected bonus amount. More importantly, users are not static responders: as Fig.~\ref{fig:user_seqential} shows, the current incentive may affect subsequent participation, incentive expectation, and session trajectory within the same interaction window. The sequentiality studied in this paper is therefore a short-horizon carryover effect. Incentivized advertising systems usually present users with a bounded number of consecutive incentive opportunities within a short window. After an incentive attempt, whether it generates an exposure can affect the user's response probability and real-time state in the next few attempts. Once the user leaves the window or the platform-side session cap is reached, the current continuous incentive process terminates.

This bounded-window structure makes incentive allocation different from both single-step response prediction and long-horizon user-lifecycle optimization. The algorithm must evaluate the immediate payoff of the current incentive while also accounting for its local effect on subsequent user states.

\section{Methodology}

\subsection{Markov Modeling}
\label{sec:markov_modeling}

We model the continuous incentive process as a Markov Decision Process (MDP). The algorithm is the agent, and users with downstream ad responses form the environment. Each transition $(s_t,a_t,r_t,s_{t+1})$ is one incentive attempt: the platform offers an amount $a_t$ in state $s_t$, then observes exposure generation, completion and payout, RTB revenue, and the next state.

\begin{figure*}[t]
  \centering
  \includegraphics[width=\textwidth]{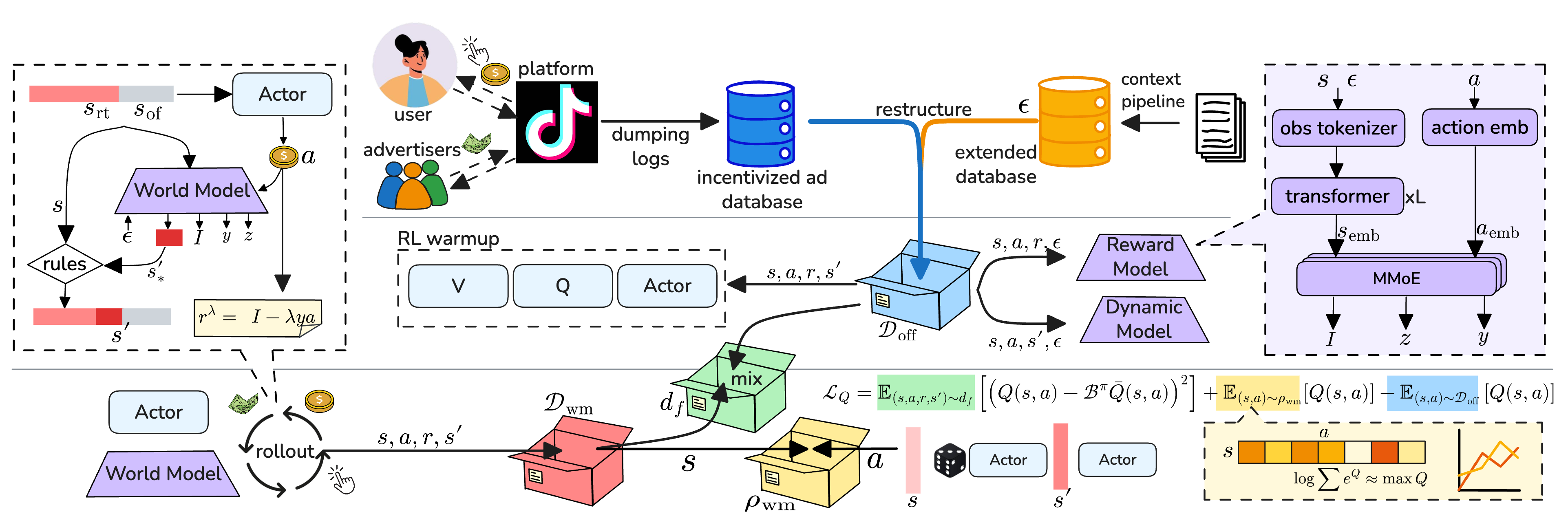}
  \caption{Offline-MBRL pipeline for incentivized advertising. (i) App logs form offline RL transitions $\mathcal{D}_{\mathrm{off}}$, with useful request context $\epsilon$ joined offline after timestamp masking for WM and CES. (ii) $\mathcal{D}_{\mathrm{off}}$ trains both the RL policy and the structured user-response WM. (iii) The Actor rolls out with the WM to generate virtual transitions, and policy optimization then uses mixed logged and virtual data with conservative regularization. In the lower panel, $d_f$ is the mixed transition distribution and $\rho_{\mathrm{wm}}$ is the WM-induced state--action proposal distribution used by the conservative term.}
  \Description{Illustration of the Offline-MBRL workflow. The upper part shows serving logs, table materialization, auxiliary pipeline joins, and initial training of a base RL policy and world model. The lower part shows policy-WM rollouts and fine-tuning with mixed real and synthetic data.}
  \label{fig:world_model}
\end{figure*}

One session-level continuous incentive interaction is an episode. Let $H_i$ be the realized number of attempts in episode $i$, and $L_{\max}$ the product-side cap, so $H_i\leq L_{\max}$. In our industrial setting, $L_{\max}$ is usually 6--20, corresponding to roughly one hour. Any attempt with no exposure terminates the episode because a product-level cooldown prevents immediate re-entry; otherwise, the interaction continues until the cap.

Because attempts are event-driven and unevenly spaced, we use a real-time half-life discount. Let $\Delta \tau_t$ be the elapsed wall-clock time between consecutive attempts and $T_{1/2}$ the half-life hyperparameter:
\begin{equation}
\gamma_t=2^{-\Delta \tau_t/T_{1/2}}
=\exp\left(-\frac{\ln 2}{T_{1/2}}\Delta \tau_t\right),
\label{eq:half_life_discount}
\end{equation}
The return of episode $i$ is $\sum_{t=0}^{H_i-1}\left(\prod_{k=0}^{t-1}\gamma_k\right)r_t^\lambda$; with fixed intervals, this reduces to standard constant-$\gamma$ notation.

The state $s_t$ contains platform-available user and context features. We split it into high-frequency real-time features $s_t^{\mathrm{rt}}$ for recent behavior, session signals, and immediate feedback, and day/week-level offline features $s_t^{\mathrm{of}}$ for user attributes, long-term activity, and historical ad response. The policy takes only $s_t$ and outputs the absolute incentive value $a_t$. Let $z_t\in\{0,1\}$ indicate exposure, $y_t\in\{0,1\}$ completion and payout, and $I_t\geq0$ the realized attempt-level RTB revenue. Completion requires exposure, so $y_t\leq z_t$; when $z_t=0$, both $y_t$ and $I_t$ are zero. The realized reward is
\begin{equation}
\label{eq:immediate_reward}
r_t^\lambda = I_t-\lambda y_t a_t,
\end{equation}
where exposure is not multiplied again because it is already absorbed into $I_t$. RTB billing does not wait for completion, and $a_t$ is not an input to the downstream RTB process. However, by changing which attempts generate exposures, $a_t$ may change the composition and conditional revenue of requests entering RTB.

\subsection{World Model Design}
\label{sec:world_model_design}

The bounded-window structure lets the WM focus on local session dynamics. Besides state $s$, offline records contain useful request context $\epsilon_t$ joined before training and materialized before $a_t$ is chosen. $\epsilon$ enters only the request-level prediction heads of the WM and CES; the policy still takes only $s$. Because $\epsilon$ is unavailable for synthetic future requests, and can't be reached by policy in online serving environments. We use one-step rollouts when $\epsilon$ is present. Appendix~\ref{app:action_invariance} details its dataflow.

We denote the learned WM by
\begin{equation}
\widehat{\mathcal{W}}_{\eta}
=\left(\widehat{R}_{\mu},\widehat{P}_{\nu}\right),
\quad \eta=\{\mu,\nu\},
\end{equation}
where $\widehat{R}_{\mu}$ and $\widehat{P}_{\nu}$ are the reward and dynamics models. The reward model has three heads:
\begin{equation}
\begin{aligned}
\hat p_{\mu,z}(s,a,\epsilon)
&\approx p(z=1\mid s,a,\epsilon),\\
\hat p_{\mu,y\mid z}(s,a,\epsilon)
&\approx p(y=1\mid z=1,s,a,\epsilon),\\
\hat m_{\mu,I}(s,a,\epsilon)
&\approx \mathbb{E}[I\mid z=1,s,a,\epsilon].
\end{aligned}
\end{equation}
Their implied one-step expected reward is
\begin{equation}
\begin{aligned}
\hat{\bar r}_{\mu}^{\lambda}(s,a,\epsilon)
&=\hat p_{\mu,z}(s,a,\epsilon)\hat m_{\mu,I}(s,a,\epsilon)\\
&\quad-\lambda \hat p_{\mu,z}(s,a,\epsilon)
\hat p_{\mu,y\mid z}(s,a,\epsilon)a.
\end{aligned}
\label{eq:wm_reward}
\end{equation}
These heads correspond to $\hat z$, $\hat y$, and $\hat I$ in Fig.~\ref{fig:world_model}. During rollout, we sample exposure and completion from the first two heads, set synthetic revenue to zero without exposure and to $\hat m_{\mu,I}$ otherwise, and compute the sampled reward by Eq.~\ref{eq:immediate_reward}; Eq.~\ref{eq:wm_reward} is its conditional expectation. The dynamics model predicts only key mutable fields in real-time user features,
\begin{equation}
\hat p_{\nu}(s'_* \mid s,a,\epsilon),
\end{equation}
where $s'_*$ denotes the mutable subvector of $s'_{\mathrm{rt}}$; for regression targets, the model may instead predict its deterministic mean $\hat{s}'_*=g_{\nu}(s,a,\epsilon)$. The full next state $s'$ combines predicted mutable fields, unchanged fields, and business rules, reducing direct prediction to the key intermediate variables.

In implementation, $\widehat{R}_{\mu}$ and $\widehat{P}_{\nu}$ are independent MMoE-based subnetworks~\cite{mmoe}. The exposure head $\hat p_{\mu,z}$ uses binary cross-entropy on all requests. The completion head $\hat p_{\mu,y\mid z}$ uses binary cross-entropy only on exposure samples ($z=1$), so it estimates conditional completion rather than an independent marginal. The conditional revenue head $\hat m_{\mu,I}$ also trains on exposure samples; alternatively, it can use ZILN loss for zero-inflated, long-tailed distributions~\cite{ziln_ltv}. The dynamics model estimates $s'_*$ with feature-type-matched losses, such as mean squared error (MSE)~\cite{bishop2006pattern} or Huber loss~\cite{huber1964robust}.

\subsection{Policy Optimization with Mixed Data}

After training the WM, we generate short rollouts from logged states. Let $\mathcal{D}_{\mathrm{off}}$ be real transitions and $\mathcal{D}_{\mathrm{wm}}$ WM-generated transitions whose sampled rewards follow Eq.~\ref{eq:immediate_reward}; Eq.~\ref{eq:wm_reward} gives their conditional expectation under the WM heads. For each synthetic transition, elapsed time is sampled uniformly from a $\pm5\%$ neighborhood of the median within-user inter-attempt interval, and the discount follows Eq.~\ref{eq:half_life_discount}. Let $d_{\mathrm{off}}$ and $d_{\mathrm{wm}}$ be the empirical transition distributions induced by these two datasets. Following the notation in Fig.~\ref{fig:world_model}, policy learning uses
\begin{equation}
d_f
=f d_{\mathrm{off}}+(1-f)d_{\mathrm{wm}},
\quad f\in[0,1],
\label{eq:mixed_transition_distribution}
\end{equation}
so the critic trains on the same mixed transition distribution $d_f$ shown in the figure. All model-based variants combine this WM augmentation with conservative Q regularization. The common starting point is the schematic objective displayed in Fig.~\ref{fig:world_model}:
\begin{equation}
\begin{aligned}
\mathcal{L}_{Q}^{\mathrm{base}}(Q)
&=
\mathbb{E}_{(s,a,r,s')\sim d_f}
\left[
\left(
Q(s,a)-\mathcal{B}^{\pi}\bar Q(s,a)
\right)^2
\right]\\
&\quad+
\mathbb{E}_{(s,a)\sim\rho_{\mathrm{wm}}}
\left[Q(s,a)\right]\\
&\quad-
\mathbb{E}_{(s,a)\sim\mathcal{D}_{\mathrm{off}}}
\left[Q(s,a)\right].
\end{aligned}
\label{eq:figure3_conservative_start}
\end{equation}
Here $\mathcal{B}^{\pi}\bar Q$ is the target-network Bellman backup. The first term learns from $d_f$, whereas the last two terms form the conservative gap: $\rho_{\mathrm{wm}}$ supplies WM-induced candidate state--action pairs and $\mathcal{D}_{\mathrm{off}}$ anchors their values to logged pairs. Equation~\ref{eq:figure3_conservative_start} uses unit weight for this gap, as in the schematic figure; the implementation below introduces a tunable coefficient.

In implementation, for each WM transition $(s,a,s')$, we construct a uniform proposal $q_u$ and noise-perturbed Actor proposals $q_s$ and $q_{s'}$. We evaluate all proposals at the current state $s$ and use their induced $\rho_{\mathrm{wm}}$ in Eq.~\ref{eq:figure3_conservative_start}. Following the official CQL implementation~\cite{cql}, we replace the positive expectation with a temperature-$T$ log-mean-exp over equally many samples:
\begin{equation}
\begin{aligned}
\mathcal{R}_{\mathrm{cons}}(Q)
&=
\mathbb{E}_{(s,a,s')\sim\mathcal{D}_{\mathrm{wm}}}
\Bigl[
\operatorname{LME}_{T}
\bigl(
Q(s,\tilde a)-\log q(\tilde a)
\bigr)
\Bigr]\\
&\quad-
\mathbb{E}_{(s,a)\sim\mathcal{D}_{\mathrm{off}}}
\left[Q(s,a)\right],
\\[-2pt]
&\hspace{32pt}
q\in\{q_u,q_s,q_{s'}\},\quad \tilde a\sim q,
\\
\mathcal{L}_{Q}^{\mathrm{MB}\mbox{-}X}
&=
\mathcal{L}_{Q}^{X}(d_f)
+
\beta\mathcal{R}_{\mathrm{cons}}(Q),
\quad
X\in\{\mathrm{TD3{+}BC},\mathrm{IQL}\}.
\end{aligned}
\label{eq:conservative_critic}
\end{equation}
Here $\mathcal{L}_{Q}^{X}(d_f)$ is the backbone-specific critic/value loss on the mixed transitions: the clipped double-Q TD loss for TD3+BC, and IQL's TD-Q and expectile-value losses. If this backbone term is the squared Bellman error, $\beta=1$, and the positive term uses the plain proposal expectation, Eq.~\ref{eq:conservative_critic} reduces to the starting form in Eq.~\ref{eq:figure3_conservative_start}. The proposal-density correction $-\log q(\tilde a)$ makes the three sampling sources comparable, and $\beta$ controls pessimism. Thus, $\rho_{\mathrm{wm}}$ and $d_f$ have distinct roles: the former identifies candidates requiring conservative control, while the latter supplies real and synthetic transitions for value learning. The penalty lowers Q-values for high-value WM candidates, while the real-data anchor prevents broad value suppression on logged pairs.

The model-based extension preserves each backbone's value and actor objectives, adding only WM transitions and the targeted conservative term; it introduces neither a planner nor a new policy architecture.

The actor updates follow their backbones. MB-TD3+BC uses the original TD3+BC policy update, whereas IQL and MB-IQL use deterministic IQL+DDPG+BC policy extraction~\cite{iqlddpg}, rather than AWR-style extraction. Both use
\begin{equation}
\begin{aligned}
\mathcal{L}_{\pi}^{\mathrm{DDPG+BC}}
&=
\mathbb{E}_{(s,a)\sim\mathcal{B}_{\mathrm{off}}}
\left[
-
\alpha
\frac{Q(s,\pi_\phi(s))}
{\operatorname{sg}\!\left(\overline{|Q|}_{\mathcal{B}}\right)+\delta}
+
\|\pi_\phi(s)-a\|_2^2
\right],\\
\overline{|Q|}_{\mathcal{B}}
&=
\frac{1}{|\mathcal{B}_{\mathrm{off}}|}
\sum_{s\in\mathcal{B}_{\mathrm{off}}}
\left|Q(s,\pi_\phi(s))\right|,
\end{aligned}
\label{eq:ddpg_bc_actor}
\end{equation}
where $\mathcal{B}_{\mathrm{off}}$ is a mini-batch of real logged pairs for the BC constraint, $\alpha$ controls policy-improvement strength, $\delta$ is for numerical stability, $\operatorname{sg}(\cdot)$ denotes stop-gradient, and $Q$ is the conservatively regularized critic. Normalization by the detached mini-batch mean absolute Q-value reduces reward-scale sensitivity and keeps actor-update scale distinct from the cost multiplier $\lambda$. Synthetic transitions improve critic learning through $d_f$, the penalty suppresses WM exploitation, and the actor BC term stays anchored to logged behavior.

\subsection{Offline Counterfactual Evaluation Scorer}

Online A/B tests are limited by traffic, incentive cost, and business risk, while offline development produces many candidate policies across algorithms, cost regimes, and training checkpoints. Rather than fitting a policy-specific evaluator for every candidate or reusing the WM optimized for policy learning, we train one independent Counterfactual Evaluation Scorer (CES) for each logged dataset. CES makes request-level counterfactual predictions on real held-out states and then aggregates them by user to score each policy in the same units as the online per-user KPIs.

Following online A/B splits, we divide logs into user-disjoint training and test sets. The policy, WM, and CES train only on training users; evaluation uses the fixed real requests of test users. CES independently estimates the same three request-level targets as the WM, denoted by $\hat p_z^{\mathrm{CES}}$, $\hat p_{y\mid z}^{\mathrm{CES}}$, and $\hat m_I^{\mathrm{CES}}$. Let $\mathcal{U}_{\mathrm{test}}$ be the held-out user set and $\mathcal{T}_u$ the requests retained for user $u$. For each request, we retain $(s_{u,t},\epsilon_{u,t})$, replace the logged incentive with $a_{u,t}^\pi=\pi(s_{u,t})$, and write $x_{u,t}^\pi=(s_{u,t},a_{u,t}^\pi,\epsilon_{u,t})$. We then sum predictions within each user and average across users:
\begin{equation}
\begin{aligned}
\widehat R_{\mathrm{CES}}(\pi)
&=
\frac{1}{|\mathcal{U}_{\mathrm{test}}|}
\sum_{u\in\mathcal{U}_{\mathrm{test}}}
\sum_{t\in\mathcal{T}_u}
\hat p_z^{\mathrm{CES}}(x_{u,t}^\pi)
\hat m_I^{\mathrm{CES}}(x_{u,t}^\pi),\\
\widehat C_{\mathrm{CES}}(\pi)
&=
\frac{1}{|\mathcal{U}_{\mathrm{test}}|}
\sum_{u\in\mathcal{U}_{\mathrm{test}}}
\sum_{t\in\mathcal{T}_u}
\hat p_z^{\mathrm{CES}}(x_{u,t}^\pi)
\hat p_{y\mid z}^{\mathrm{CES}}(x_{u,t}^\pi)
a_{u,t}^\pi,\\
\widehat J_{\mathrm{CES}}(\pi)
&=
\widehat R_{\mathrm{CES}}(\pi)
-\lambda_{\mathrm{eval}}\widehat C_{\mathrm{CES}}(\pi).
\end{aligned}
\label{eq:ces_score}
\end{equation}

Here $\widehat R_{\mathrm{CES}}$ and $\widehat C_{\mathrm{CES}}$ are estimated revenue per user (revenue/u.) and incentive cost per user (cost/u.) over the evaluation window. Every held-out user receives equal weight, while that user's request-level predictions are summed. The weight $\lambda_{\mathrm{eval}}=1$ therefore makes $\widehat J_{\mathrm{CES}}$ estimated net profit per user (net p./u.); another business weight may be used when matching a training objective. This user-level score still conditions on the same fixed logged requests. It is not full long-horizon OPE and does not let candidate policies recursively alter later state or request-count distributions.

Within this scope, a dataset-specific CES is reusable across policies: after held-out user requests are materialized, each checkpoint requires only batched Actor and CES inference followed by user-level aggregation. This makes the evaluation loop resemble supervised validation and supports frequent training diagnostics, while preserving real users, requests, context, and traffic composition. CES remains a support-aware pre-launch ranking tool rather than an unbiased estimator of absolute online KPIs. Appendix~\ref{app:evaluation_efficiency} reports its supplementary timing comparison with FQE under a fixed GPU budget.

\section{Experiment}

We evaluate cost sensitivity, compare policies through CES and online A/B tests, and assess CES and WM reliability.

We use three large-scale logged environments from different serving periods, app platforms, target-user groups, and behavior policies: DY-1($\sim$200M), DY-2($\sim$400M), and DH($\sim$50M). DY-2 and the subsequent online experiments use the same environment, so our main analysis uses DY-2. For each environment, we train on its training users and evaluate on its test users. The split follows online A/B testing: all records for a user belong to either training or test. Appendix~\ref{app:exp_config} details the data sources and shared training settings.

\subsection{Cost Tendency and ROI Preference}
\label{sec:cost_tendency}

\begin{figure}[t]
  \centering
  \includegraphics[width=\linewidth]{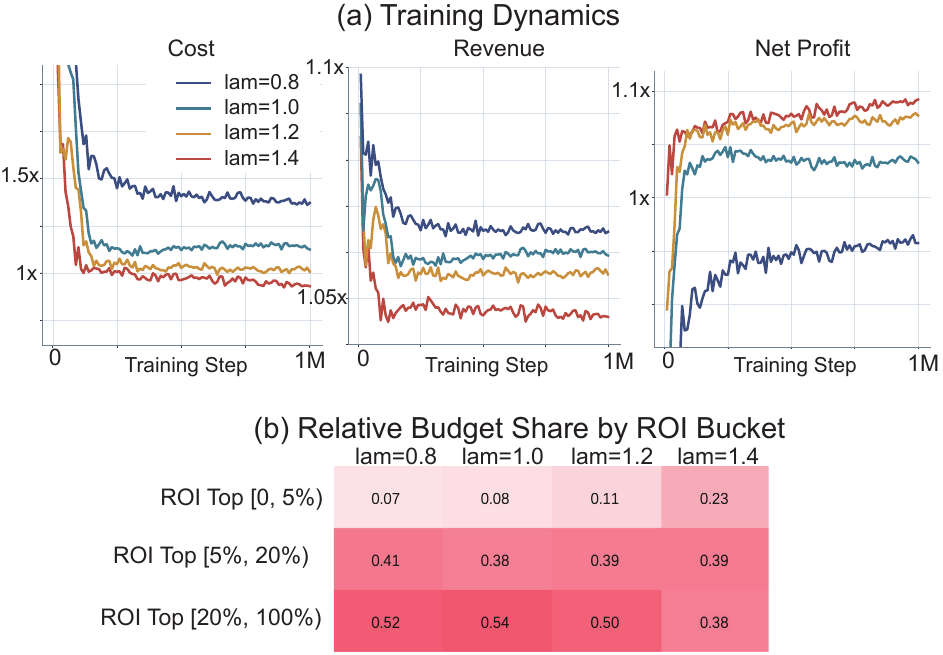}
  \caption{Increasing $\lambda$ makes the learned policy more cost-sensitive and shifts budget toward high-ROI users. (a) relative cost, revenue, and net profit during training; (b) relative budget share across ROI buckets.}
  \Description{The left panel shows that increasing the cost coefficient reduces cost and revenue while improving net profit in DY-2. The right panel shows that higher cost coefficients allocate a larger share of budget to high-ROI users.}
  \label{fig:cost_control}
\end{figure}

The realized reward $r_t^\lambda=I_t-\lambda y_ta_t$ uses a fixed-multiplier Lagrangian scalarization. In expectation, it trades advertiser revenue against completed-incentive payout. Each value of $\lambda$ therefore defines a cost-preference regime. A larger $\lambda$ makes the policy more cost-sensitive and favors opportunities with clearer marginal net gains. We study this relation on DY-2, the environment used in the subsequent online A/B tests. As Fig.~\ref{fig:cost_control}(a) shows, increasing $\lambda$ decreases both cost and revenue while increasing net profit in this environment. Fig.~\ref{fig:cost_control}(b) further shows the corresponding allocation effect: larger $\lambda$ shifts budget toward high-ROI users. Thus, $\lambda$ changes both total cost tendency and the policy's ROI preference, rather than acting only as a global bid scale.

Although adjusting the cost level can affect net profit, the policy-comparison experiments control overall incentive cost. This is because incentive spending can itself change long-term user metrics such as LT and LTV, which are outside the short-term net-profit objective studied here. Controlling total cost therefore isolates the policy's short-term allocation effect from changes induced merely by spending more or less. Thus, algorithmic performance here fundamentally measures the quality of incentive-budget reallocation under controlled cost. Budget ceilings are handled by outer deployment controls around screened policies rather than by changing the comparison target; Appendix~\ref{app:cost_interface} summarizes this interface.
\subsection{CES Evaluation}
\label{sec:ces_eval}

We report CES evaluation after 1M training steps for model-free RL and its model-based counterparts. The model-free backbones are TD3+BC~\cite{td3bc}, IQL~\cite{iql}, and CQL~\cite{cql}. MB TD3+BC and MB IQL augment the corresponding backbone with our structured WM; MB CQL combines the COMBO~\cite{combo} objective with the same WM design. Appendix~\ref{app:offline_reward_full} adds imitation-learning and sequence-modeling baselines, seed-level variation, and cost diagnostics.

\input{tables/offline_reward}

Using the same CES, held-out traffic, and cost protocol, every WM variant improves its matched model-free backbone. Thus, modeling local user feedback and advertising outcomes adds training signal beyond value learning alone.
\subsection{Online A/B Test}
\label{sec:online_ab}

\input{tables/online_ab}

The sequential A/B tests establish the matched-cost online ordering DESCN $<$ TD3+BC $\approx$ IQL $<$ MB IQL, consistent with CES. The successful model-based upgrade and the loss of its gain when WM is removed isolate the WM contribution beyond backbone choice. Expanded metrics show no significant LT degradation, while time-aligned and common-reference backtests recover the same deployment signs and ordering (Appendices~\ref{app:online_ab_metrics}, \ref{app:ope_ab_backtest}, and~\ref{app:ope_common_reference}). All A/B tests have 10\% peak traffic and over 10M cumulative hit users.

\subsection{CES Reliability}
\label{sec:ces_reliability}
We assess CES through factual calibration and support sensitivity. Appendix~\ref{app:action_invariance} finds little residual association between request context $\epsilon$ and the logged incentive after conditioning on $s$ (maximum correlation $0.066$; $R^2$ $0.7133\rightarrow0.7294$), supporting the use of $\epsilon$ in WM and CES.

Table~\ref{tab:ces_reliability} compares policy-visible state $s$ with context-augmented $(s,\epsilon)$ on logged outcomes and then stresses support shift. DY-2$_{\mathrm{in}}$ retains the central $\pm20\%$ of DY-2's $\pm50\%$ action-exploration band and truncates a trajectory at its first exit; DY-2$_{\mathrm{out}}$ is the held-out complement. Context improves factual discrimination and conditional-revenue prediction while keeping ECE small. In contrast, a CES trained on DY-2$_{\mathrm{in}}$ shows worse AUC, ECE, and revenue error on the full distribution and further outside its support. CES rankings are therefore locally trustworthy, and policy hyperparameters must limit action drift.

\input{tables/ces_reliability}

Appendix~\ref{app:policy_ood} complements this factual diagnostic with cross-support policy checks that delimit the regime in which CES rankings remain reliable.

\subsection{WM Reliability}
\label{sec:wm_reliability}

Table~\ref{tab:wm_reliability} compares one-step WM prediction with and without observed request context $\epsilon$, and autoregressive rollout on $s$ alone. Let $s'_{\mathrm{c}}$ and $s'_{\mathrm{b}}$ denote the mutable CPM-type and user-behavior-type field groups in $s'_*$. Conditioning on $\epsilon$ improves response gates, conditional revenue, and next-state prediction; removing it and recursively feeding generated states back into the WM amplifies error. This supports using local one-step rollouts when request context is available.

\input{tables/wm_reliability}
On the prediction targets shared with CES, the WM is less accurate because it must jointly fit a substantially heavier multitask objective: in addition to response and revenue, the $s'_{\mathrm{c}}$ and $s'_{\mathrm{b}}$ groups together contribute nearly ten prediction heads. This remains a demanding task even though the WM, unlike CES, uses an MMoE architecture to improve multitask performance.

To investigate whether conservative Q regularization (Q reg.) in Eq.~\ref{eq:conservative_critic} is always beneficial, Table~\ref{tab:q_regularization_ablation} presents an ablation matrix after 1M training steps on DY-2. Because the model-free backbones use only the logged-data distribution, we apply Q reg. directly to it. We also remove Q reg. from the model-based methods to assess its effect. Because CQL already includes its own Q regularization, we do not repeat this ablation for CQL. For TD3+BC and IQL, Q reg. slightly degrades the model-free backbones but preserves the performance of their model-based variants. Without Q reg., the latter suffer substantial performance degradation. Some entries are taken from Table~\ref{tab:offline_reward_full}, and the cost changes in the additional experiments remain within the range observed in that table.

\input{tables/q_regularization_ablation}

\textbf{Additional checks.}
Further reference experiments support the sequential formulation beyond the main matched-backbone comparison. Half-life-discounted policies score higher than one-step variants offline, and TD3+BC improves net profit over one-step causal or bandit online policies in two reference A/B tests (Appendix~\ref{app:one_step_reference}). On D4RL HalfCheetah, the public-benchmark experiments provide a supplementary viability check for the model-based extensions under a standard transition-and-reward setup (Appendix~\ref{app:d4rl}).

\section{Related Work}

\textbf{Incentivized advertising and auto-bidding.}
Incentive mechanisms have appeared in several advertising settings. Incentivized social advertising~\cite{incentivized_social_ads} pays influential users to endorse or propagate promoted content, focusing on seed selection and budget allocation under social diffusion constraints. This confirms that incentives can be part of an advertising mechanism, but the decision object is social propagation rather than platform-side ad-watching incentives before a downstream monetization pipeline is triggered.
The closest advertising literature is auto-bidding in RTB. RL-based bidding methods formulate repeated bidding under budget constraints as sequential decision making~\cite{rtb_rl,drlb}, and recent generative approaches such as AIGB~\cite{aigb} and GAVE~\cite{gave} model bid trajectories from return or value signals. These methods make sequential, budget-sensitive advertising decisions at industrial scale, but they operate after an exposure already exists. Our problem is upstream: the platform chooses a user incentive before knowing whether an exposure will be generated or what ad revenue it will produce.

\textbf{Targeted promotion and uplift-based incentive allocation.}
Targeted promotion studies how to allocate coupons, cash bonuses, discounts, or other treatments under cost constraints. Industrial systems often follow a predict-then-optimize pattern: estimate user response or uplift, solve a constrained allocation problem, and, in some deployments, use online feedback control to respect budgets~\cite{spending_wisely,msba_meal_delivery,online_bonus_cuid,realtime_coupon_allocation}. Related academic formulations add structured counterfactual inference, differentiable allocation, or offline RL to this pipeline~\cite{ccposire,e3ir,bcrlsp}. These studies are close in spirit because they treat incentives as costly actions in logged data. Their objectives, however, are usually conversion, acquisition, order acceptance, retention, or LTV over existing user demand. In incentivized advertising, the incentive first determines whether an exposure is generated. If no exposure is generated, no ad revenue follows. The reward must therefore combine user response, RTB revenue, and incentive payout, and the policy must account for short-window carryover before deployment.


\textbf{Offline and model-based RL.}
Offline RL learns policies from fixed logged datasets without further environment interaction. Its central challenge is distribution shift: learned policies may choose actions poorly covered by the behavior data, causing extrapolation error and value overestimation. Representative methods use conservative value learning, behavior regularization, or implicit policy improvement, including CQL~\cite{cql}, TD3+BC~\cite{td3bc}, IQL~\cite{iql}, and DT~\cite{decision_transformer}. These methods provide the main model-free baselines in our experiments. Offline Model-based RL learns a world model and uses simulated transitions to improve policy learning. MOReL~\cite{morel}, MOPO~\cite{mopo}, COMBO~\cite{combo}, and LEQ~\cite{leq} limit model exploitation through pessimism, uncertainty penalties, or conservative value regularization. These approaches are not directly tailored to incentive allocation, where the horizon is bounded, returns combine stochastic response gates with posterior RTB revenue, and request-level context is available offline to the WM but unavailable to the deployed policy or future synthetic states; consequently, long multi-step rollouts can compound context-induced model bias. Our method therefore uses a business-structured WM for local counterfactual incentive optimization and conservative value control, rather than generic long-horizon user simulation.

\section{Conclusion}

This paper studies incentive allocation in incentivized advertising, where the platform promises incentives before observing downstream ad revenue. We formulate this setting as a cost-sensitive sequential decision problem and propose an Offline-MBRL framework that learns structured user-response and revenue models from logged data, augments conservative policy optimization with local WM rollouts, and uses an independent CES to screen candidate policies before launch. Across industrial offline environments, CES consistently ranks the model-based policies above their corresponding model-free counterparts. Online A/B tests confirm this deployment ranking, and backtests further corroborate it. These results show that modeling user feedback and ad revenue can improve incentive allocation without requiring risky online RL exploration.

\noindent\textbf{Limitations.} Our formulation targets short-horizon interaction windows and optimizes net profit; longer-term user outcomes are monitored online but are not explicit optimization targets. CES is deliberately a support-aware, user-level ranking tool built from request-level predictions, rather than a general estimator of full long-horizon counterfactual value. Moreover, the WM and CES overlap substantially in their inputs and prediction targets, so their errors may be correlated, creating a risk of overestimating policy value. Although the online A/B tests confirm the incremental gains from WM augmentation, developing efficient offline evaluation methods that are independent of the policy-learning model remains an important direction for future work.

\bibliographystyle{ACM-Reference-Format}
\bibliography{refs_orl,refs_business}

\appendix

\section{Cost-Sensitive Control and Deployment Interface}
\label{app:cost_interface}

The main text formulates incentive allocation as a cost-sensitive sequential decision problem, characterizes the cost tendency induced by $\lambda$, and compares policies under controlled cost. This appendix clarifies the cost-alignment criteria and multi-layer cost control used in deployment.

\subsection{Cost Alignment and Long-Term Metrics}

The primary offline and online comparisons control overall incentive cost so that policy gains reflect allocation rather than a change in total spending. Online, matched cost means that the difference in realized per-user incentive cost is not significant at the $p=0.05$ level. Offline, we use a common factual-cost alignment protocol for CES screening. For each algorithm $m$, we train candidates with $\lambda\in\{0.5,0.6,\ldots,2.0\}$ and select, on the same held-out users and requests, the candidate whose CES-estimated cost/u. is closest to the factual cost/u. of the logged actions:
\begin{equation}
\pi_m^*=\arg\min_{\pi\in\Pi_m}
\left|\widehat C_{\mathrm{CES}}(\pi)/C_{\mathrm{fact}}-1\right|.
\label{eq:offline_cost_alignment}
\end{equation}
Both $\widehat C_{\mathrm{CES}}(\pi)$ and $C_{\mathrm{fact}}$ use the sum-within-user, average-across-users convention of Eq.~\ref{eq:ces_score}. Selection uses cost proximity only, after which the selected candidate is used to report CES net p./u.; the residual cost/u. changes are reported in Table~\ref{tab:offline_reward_full}. This shared reference keeps all algorithms in the delivery-cost neighborhood represented by the logged environment while avoiding unnecessary sensitivity to small scorer and traffic fluctuations. Offline comparable cost therefore denotes factual-cost-aligned screening, whereas the online A/B tests provide the realized matched-cost check. Together, these conventions isolate the short-term allocation effect while long-term metrics such as LT and LTV are monitored separately online.


\subsection{Multi-Layer Cost Control}

Cost control is implemented at three complementary layers. First, the policy layer trains and screens a family of $\lambda$-conditioned cost preference policies using Offline-MBRL and CES. Second, the candidate selection layer chooses policies under comparable-cost and ROI requirements, which provides an offline guardrail before traffic exposure. Third, the production delivery layer schedules the selected policy or $\lambda$ regime according to budget ceilings, campaign cycles, and real-time risk constraints. This division keeps the world model and conservative policy optimization focused on user response and within-window dynamics, while allowing business-side controls to be adjusted, rolled back, and audited without retraining the policy. Consequently, budget control does not rely on frequent online exploration: it is handled through offline policy screening and outer scheduling around a stable set of candidate policies.

\section{Additional Ablations}
\label{app:additional_ablations}

This appendix collects experiments that do not have a direct counterpart in the main experimental narrative. They provide additional operational reference and diagnostic evidence, rather than extending the main policy-comparison claims.

\subsection{Offline Backtesting Against Deployment Evidence}
\label{app:ope_ab_backtest}

\subsubsection{Time-Aligned Backtest}

We conduct offline backtesting on the strategy pairs confirmed by the online A/B tests in Table~\ref{tab:online_ab}. We take the original policy weights from A/B testing and mark the specific weights with subscripts, such as IQL$_B$, representing the IQL strategy weight that first appeared in Experiment B. Specifically, the DESCN weight file in Experiment A was accidentally lost, so we replaced it with a strategy distilled from its serving record and marked it as distilled with DESCN $'_A $. For the reversal, IQL$_D$ uses exactly the hyperparameters of the contemporaneous plain-IQL counterpart IQL$_B$, but is retrained on newer logs collected under MB IQL$_C$; using IQL$_D$ rather than reusing IQL$_B$ makes the test conservative with respect to a generic ``newer data is better'' explanation. All results are evaluated after the model training data deadline and before the actual A/B test deadline, aligning the offline and online environments as closely as possible.

Table~\ref{tab:ope_ab_backtest} reports cost/u. and net p./u. for the online A/B test, CES, and a separately trained fitted-Q evaluation (FQE)~\cite{fqe_batch_policy}. FQE uses only the policy-visible state, whereas CES evaluates the user-level KPI objective in Eq.~\ref{eq:ces_score} by aggregating request-level counterfactual predictions. 

\input{tables/ope_ab_backtest}

This is deliberately a pairwise backtest, rather than a claim that either score is an unbiased estimate of online KPIs. The A/B-test block gives the deployment evidence against which the signs of the offline differences can be assessed. Both CES and FQE recover the signs of the four deployment decisions.

\subsubsection{Common-Reference Backtest}
\label{app:ope_common_reference}

The time-aligned backtests above test four actual deployment decisions, but their serving regimes differ. As a complementary reference only, Table~\ref{tab:ope_common_reference} evaluates all four deployed policy checkpoints on DY-2, using the same users and requests. This removes cross-slice data variation and makes the offline ordering more comparable. 

\input{tables/ope_common_reference}

The offline scores of IQL$_B$ and IQL$_D$ are close; on the common DY-2 reference, CES likewise places IQL$_B$ and IQL$_D$ at $+2.5\%$ and $+2.1\%$, and chaining Experiments C and D returns IQL to within about 1\% of TD3+BC both online and in CES. Together, these checks reduce the plausibility of a data-recency explanation and reinforce the deployment ordering DESCN $<$ TD3+BC $\sim$ IQL $<$ MB IQL.

\subsection{One-Step and Sequential Policies}
\label{app:one_step_reference}

The following comparisons are included as reference results. 

\subsubsection{CES and FQE Results}

Table~\ref{tab:one_step_ces} compares one-step ($\gamma=0$) and half-life-discounted variants of TD3+BC and IQL. All policies are scored on the same environment. 

\input{tables/one_step_ces}

\subsubsection{Online A/B Tests}

Table~\ref{tab:one_step_ab} reports two online A/B comparisons between TD3+BC and policies that use one-step causal or bandit objectives. 

\input{tables/one_step_ab}

\subsection{Fixed-Budget Evaluation Overhead}
\label{app:evaluation_efficiency}

We supplement the accuracy-oriented backtests with a wall-time comparison of continuous checkpoint evaluation. All three settings use the same PyTorch training job and the same fixed allocation of 8 GPUs, each with 32 GB of GPU memory and approximately 16 TFLOPS of FP32 compute: (i) policy training without test evaluation, (ii) policy training with CES evaluation at each saved checkpoint, and (iii) policy training with FQE at the same checkpoints. Evaluation is synchronous and uses the same eight GPUs; no additional asynchronous workers are provisioned. This fixed job-level budget measures evaluation overhead without exchanging it for an extra compute pool.

For CES, Ray distributes a test set materialized once before policy training, and PyTorch DDP runs Actor and CES inference with a per-GPU batch size of 2048 (global batch size 16,384). The dataset-specific CES is trained once and shared by all policy checkpoints; the timed region therefore measures its repeated-use workflow. FQE is target-policy-specific, so a separate evaluator is fitted for each checkpoint using the same protocol as in Section~\ref{app:ope_ab_backtest}. We report total wall time from the first policy update to job completion and $\mathrm{Overhead}=(T_{\mathrm{eval}}-T_{\mathrm{train}})/T_{\mathrm{train}}$.

\input{tables/evaluation_efficiency}

Under the fixed GPU budget, policy training alone takes 20 hours. Adding CES increases the total wall time to 21 hours, corresponding to only 5\% overhead, whereas adding FQE increases it to 30 hours, or 50\% overhead. The CES result excludes its approximately 1-hour training cold start because the dataset-specific scorer is trained only once per dataset and then reused across policy runs and checkpoints. In this repeated-evaluation setting, large inference batches and a pre-materialized test set make offline Actor--CES inference highly efficient, leaving little incremental cost per checkpoint. By contrast, FQE must fit a new target-policy-specific evaluator for every checkpoint. This repeated optimization is substantially slower under the same fixed compute budget, especially on GPUs with 32 GB of GPU memory and approximately 16 TFLOPS of FP32 compute.

\section{Supplementary Results for Main Experiments}
\label{app:supp_results}

The main experimental section establishes the offline, online, and reliability results. The following subsections supply their detailed counterparts in the same order: complete CES offline metrics with variability, expanded online KPIs, CES support diagnostics, and WM rollout diagnostics. Together they show that the offline policy ranking is evaluated at comparable cost, is consistent with matched-cost online evidence, and is interpreted within the scorer's and world model's supported regime.

\subsection{CES Offline Evaluation: Full Baselines and Cost Diagnostics}
\label{app:offline_reward_full}

Table~\ref{tab:offline_reward_full} keeps the full offline comparison used as supporting evidence for Table~\ref{tab:offline_reward}. The main text focuses on matched model-free versus +WM comparisons, while this appendix reports imitation-learning and sequence-modeling baselines, CES-Fact. checks, cost changes, and seed-level standard deviations.

\input{tables/offline_reward_full}

\subsection{Main Online A/B Test: Detailed Setup, Metrics and More Analysis}
\label{app:online_ab_metrics}

This subsection expands the key online metrics reported in Section~\ref{sec:online_ab} by reporting detailed setup, full metrics and analysis.

\input{tables/online_ab_expanded}

\subsubsection{Setup.}
All values in Table~\ref{tab:online_ab_expanded} are relative changes of the candidate policy against the online policy served concurrently in the same experiment. The reported lifts are CUPED-adjusted estimates. Statistical significance is assessed on CUPED-adjusted DID-level outcomes using two-sided Welch $t$-tests, and the reported $p$-values incorporate the experimentation platform's standard multiplicity adjustment. The table reports per-user incentive attempts, exposures, incentive cost, advertiser revenue, net profit, incentive ROI, 7-day LT, 30-day LT, and cumulative LTV. Each group in every experiment contains more than 10 million unique device identifiers (DIDs). The four A/B tests were conducted sequentially, with no time overlap. For each experiment, traffic in each group was first ramped from 0.1\% to 5\%. After the monitored metrics remained normal at 5\%, the group was further ramped to 10\% and held at that level for measurement. At launch, eligible traffic was assigned to groups at the DID level. We used no special user segmentation or targeting rules for the split. In the reversal Experiment D, plain IQL also employs a strategy extraction method in the style of DDPG+BC, as in Experiment C.

\subsubsection{Analysis.}
We discuss only statistically significant changes in Table~\ref{tab:online_ab_expanded}. The general rollout criterion is deliberately based on business outcomes rather than delivery diagnostics: a candidate should not increase incentive cost, should significantly improve net profit, and should not degrade LT. Per-user incentive attempts, ad exposures, and incentive ROI are useful for diagnosing why a policy changes the business outcome, but they are not rollout gates on their own.

\textbf{Experiment A} satisfies this criterion. At statistically matched cost, TD3+BC significantly increases per-user net profit by 7.72\%; its ROI also improves significantly. The significant increases in incentive attempts (+1.31\%) and exposures (+1.44\%) indicate that the policy creates more incentive--advertising interactions and raises overall advertising penetration. Thus, the profit gain is achieved without an incremental-cost or LT concern, supporting full rollout.

\textbf{Experiment B} has matched cost and net profit, but significantly fewer incentive attempts (-1.10\%) and exposures (-1.22\%). This pattern is consistent with a post-hoc allocation explanation: relative to the TD3+BC policy in Experiment A, IQL appears to direct more cost toward a small set of higher-eCPM users. Although these users are too few to sustain aggregate delivery and exposure volume, each can contribute meaningful advertising value. Nevertheless, this is a reallocation at flat net profit rather than a demonstrated business improvement; therefore, it does not satisfy the rollout criterion.

\textbf{Experiment C} again satisfies the criterion. MB IQL keeps cost statistically unchanged while significantly increasing revenue by 6.80\% and net profit by 7.96\%. No monitored LT metric changes significantly, so the matched-cost revenue and net-profit gains provide the evidence needed for rollout without an LT concern.

\textbf{Experiment D} is a reversal test that illustrates the opposite allocation pattern. Relative to the MB IQL baseline from Experiment C, plain IQL keeps cost statistically unchanged but significantly decreases revenue (-4.77\%), net profit (-6.56\%), and ROI (-5.32\%), while incentive attempts (+21.81\%) and exposures (+23.68\%) increase sharply. This combination is consistent with allocating more incentive cost to a broad low-eCPM population: the larger population contributes many more incentive interactions and exposures, but its lower underlying advertising value reduces realized revenue. Consequently, more exposures do not translate into more value, and the reduction in revenue leads to lower net profit and ROI. This policy should therefore not be rolled out.

More broadly, these delivery and ROI metrics are post-hoc explanatory signals, not direct learning targets. The balance between high- and low-eCPM users, the recipients of incentives, and the number of incentives delivered are all implicit consequences of optimizing net profit under matched cost. We do not manually prescribe these allocation choices. Finally, all four experiments have statistically matched cost, and none shows a significant change in the monitored LT metrics; within the sensitivity of these tests, the policy changes therefore do not produce a detectable long-term impact.

\subsection{Further Discussion of Context Enrichment and CES Support}
\label{app:policy_ood}

We first document the dataflow and action-invariance of enriched request context, then test whether policy OOD changes CES evaluation. These diagnostics characterize the support within which the CES ranking signal is used.

\subsubsection{Context Dataflow}

\begin{figure}[t]
  \centering
  \includegraphics[width=\linewidth]{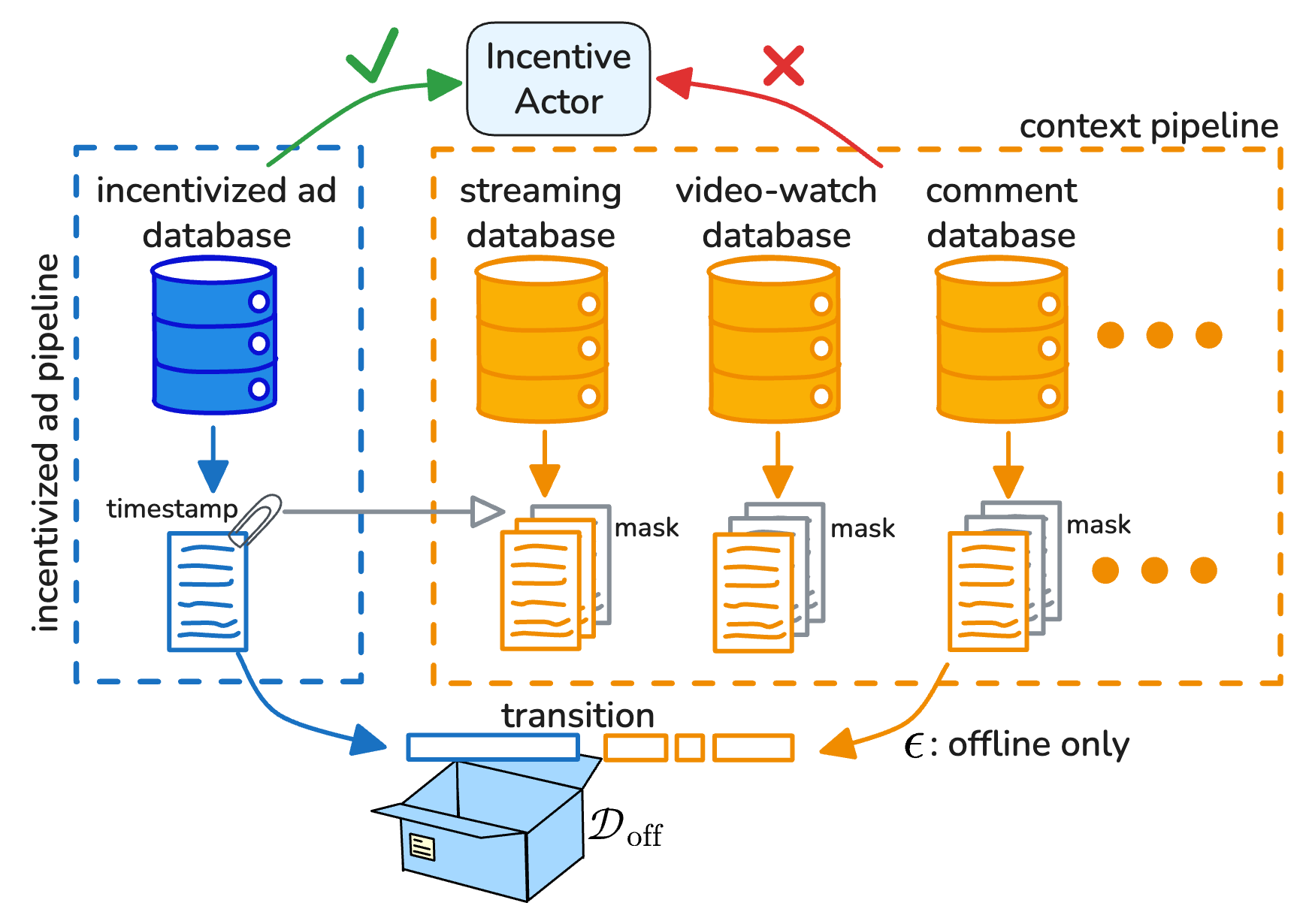}
  \caption{Timestamp masking prevents future-information leakage when constructing offline request context $\epsilon_t$. The timestamp of each incentivized-ad record masks auxiliary context records that would not yet have been available at that request. The masked context is retained only in the offline transition dataset $\mathcal{D}_{\mathrm{off}}$ to assist the WM and CES; the online Incentive Actor cannot access $\epsilon_t$ because of the serving architecture.}
  \Description{A pipeline schematic with an incentivized-ad database on the left and several auxiliary context databases on the right.  The timestamp from the incentivized-ad record is used to mask the context records before they are combined into an offline transition dataset.  The Incentive Actor receives the serving-state path from the incentivized-ad pipeline, while a crossed-out arrow indicates that it cannot receive the offline-only enriched context epsilon.}
  \label{fig:context_pipeline}
\end{figure}

Figure~\ref{fig:context_pipeline} is a pipeline schematic. For each incentivized-ad record at time $t$, we use the record timestamp to mask every joined context source, retaining only information that was available at or before $t$.  This prevents future streaming, video-watch, comment, or other auxiliary-pipeline records from leaking post-decision or post-outcome information into the enriched request context $\epsilon_t$.

The resulting $\epsilon_t$ is materialized only in the offline dataset $\mathcal{D}_{\mathrm{off}}$. It is used as an auxiliary input for the request-level prediction heads of the WM and CES, where it improves factual response and revenue estimates on logged requests; CES subsequently aggregates those predictions by user. In contrast, the deployed Incentive Actor receives only the policy-visible serving state $s_t$ from the incentivized-ad pipeline: it neither reads $\epsilon_t$ nor can access the auxiliary context pipelines at serving time. Thus, context enrichment supports offline model learning and policy evaluation without creating an online serving dependency or exposing the policy to unavailable information.

\subsubsection{Action-Invariance of Request Context}
\label{app:action_invariance}

\begin{figure}[t]
  \centering
  \includegraphics[width=\linewidth]{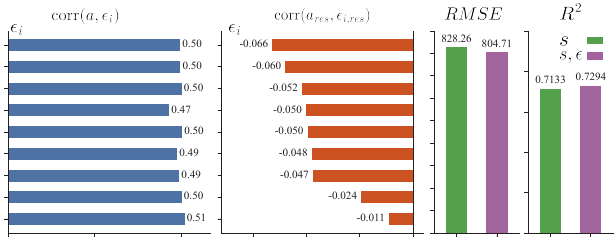}
  \caption{CES action-invariance diagnostics. Raw $a$--$\epsilon$ correlations are state-driven; after residualizing on $s$, $|\mathrm{corr}|<0.1$ (max $0.066$, mean $0.045$). Predicting $a$ improves only modestly from $s$ to $(s,\epsilon)$ (RMSE $828.26\to804.71$, $R^2$ $0.7133\to0.7294$).}
  \Description{A four-panel diagnostic figure. The first panel shows raw correlations between logged incentives and ranking-side request-context features around 0.47 to 0.51. The second panel shows residual correlations after conditioning on state, all within plus or minus 0.1. The last two panels compare logged-incentive prediction from state only and from state plus request context, showing a modest RMSE reduction and R-squared increase.}
  \label{fig:ces_residual_corr}
\end{figure}

The enriched request context $\epsilon$ is joined from surrounding serving systems and is excluded from the online policy input. Its source pipelines do not take the platform incentive as input. We further check that $\epsilon$ carries little incremental information about the logged incentive beyond the policy-visible state by residualizing both variables on $s$:
\begin{equation}
a_{\mathrm{res}}=a-\widehat{\mathbb{E}}[a\mid s],
\quad
\epsilon_{i,\mathrm{res}}=\epsilon_i-\widehat{\mathbb{E}}[\epsilon_i\mid s].
\end{equation}
As shown in Fig.~\ref{fig:ces_residual_corr}, raw correlations are mostly explained by state: after residualization, all absolute correlations are below $0.1$. Predicting logged $a$ from $(s,\epsilon)$ only modestly improves over using $s$ alone. These results support the request-context design used by CES; features that fail this diagnostic are excluded from CES and used only for WM-side auxiliary training or factual analysis.

These results suggest that $\epsilon_t$ is best viewed as decision-time environment information that objectively affects the transition but is inaccessible to the deployed policy. It can nevertheless benefit critic-side learning indirectly through $\epsilon$-conditioned reward estimates and WM-generated transitions, without becoming an Actor input. This asymmetry admits a partially observable MDP interpretation, with $s_t$ as the policy observation and $\epsilon_t$ as additional environment context. We deliberately do not formalize observability in this work, however, because doing so would require resolving whether the next context $\epsilon_{t+1}$ is predictable from the available history and, if it is not, whether the model state should be augmented with further information that makes $\epsilon_{t+1}$ predictable. These questions are orthogonal to our focus on structured one-step modeling and supported offline policy evaluation.

\subsubsection{Cross-Support CES Evaluation}

We train three policies with progressively lower delivery on DY-2$_{\mathrm{in}}$: BC, IQL ($\lambda=1.5$), and the deliberately extreme IQL ($\lambda=8.0$). Each checkpoint is scored by $\mathrm{CES}_{\mathrm{in}}$, trained on DY-2$_{\mathrm{in}}$, and $\mathrm{CES}_{\mathrm{all}}$, trained on full DY-2. This comparison isolates evaluator bias as a policy moves beyond the inner support.

\begin{figure}[t]
  \centering
  \includegraphics[width=\linewidth]{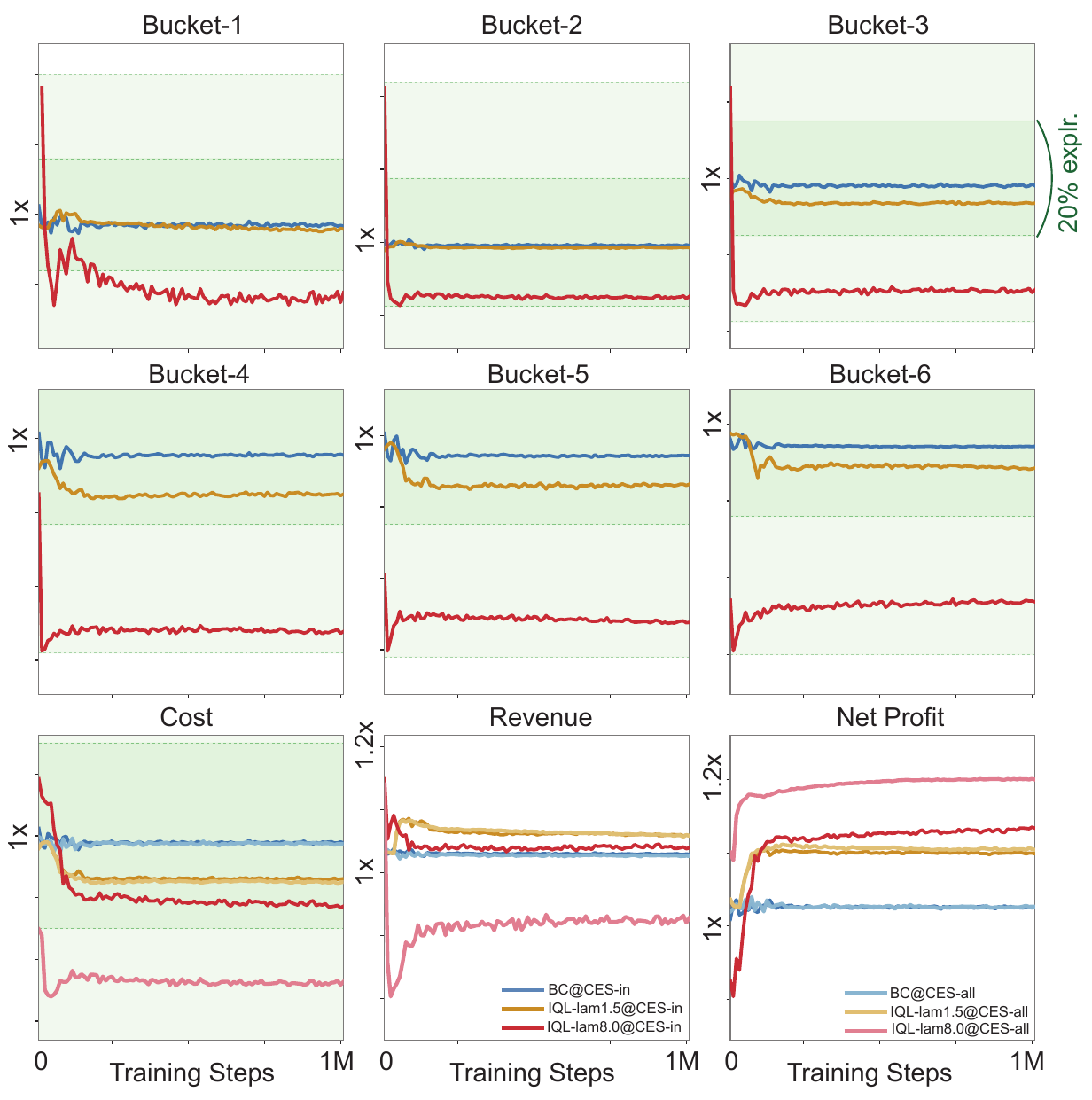}
  \caption{Cross-support CES diagnostic. Top: mean actions in six eCPM buckets, with $\pm20\%$ DY-2$_{\mathrm{in}}$ (dark green) and $\pm50\%$ DY-2 (light green) supports; cost/u. carries the corresponding per-user payout ranges. Bottom: user-aggregated $\mathrm{CES}_{\mathrm{in}}$ (dark) and $\mathrm{CES}_{\mathrm{all}}$ (light) agree for BC and IQL ($\lambda=1.5$), but diverge for IQL ($\lambda=8.0$) outside the inner support.}
  \Description{A three-by-three diagnostic figure. The first six panels show BC, IQL with lambda 1.5, and IQL with lambda 8.0 across six eCPM buckets and two nested action-support bands. The bottom panels compare cost/u., revenue/u., and net p./u. estimated by CES-in and CES-all. Estimates overlap for BC and IQL with lambda 1.5; for IQL with lambda 8.0, CES-in estimates higher cost/u. and revenue/u. but lower net p./u. than CES-all.}
  \label{fig:action_drift}
\end{figure}

We partition users into six equal-width eCPM buckets and track the mean action in each bucket. Let $a_i^0$ be the behavior policy's pre-exploration action for transition $i$, and let $\mathcal{I}_u$ index the fixed test transitions of user $u$. Define
\begin{equation}
\mathcal{A}_{i,\rho}
=
\left[a_{i,\rho}^{-},a_{i,\rho}^{+}\right]
=
\left[(1-\rho)a_i^0,(1+\rho)a_i^0\right],
\qquad
\rho\in\{0.2,0.5\}.
\label{eq:cross_support_action}
\end{equation}
The dark- and light-green bands in Fig.~\ref{fig:action_drift} mark the DY-2$_{\mathrm{in}}$ ($\rho=0.2$) and full DY-2 ($\rho=0.5$) supports. BC and IQL ($\lambda=1.5$) stay inside the inner band in all six buckets. IQL ($\lambda=8.0$) repeatedly enters the gap between the bands in every bucket, but remains inside the full support.

The cost/u. panel carries the same distinction through the corresponding per-user payout ranges. Revenue/u. and net p./u. do not inherit the action percentages because the action changes the response gates and the selected RTB-request composition. For each fixed test transition, let
\[
m_{i,\rho}^{-}=\min_{a\in\mathcal{A}_{i,\rho}}\hat m_I^{\mathrm{CES}}(s_i,a,\epsilon_i),
\qquad
m_{i,\rho}^{+}=\max_{a\in\mathcal{A}_{i,\rho}}\hat m_I^{\mathrm{CES}}(s_i,a,\epsilon_i).
\]
An outcome-extreme envelope induced by $\mathcal{A}_{i,\rho}$ is
\begin{equation}
\begin{aligned}
0\leq \mathrm{Rev}_{\rho}
&\leq
\frac{1}{|\mathcal{U}_{\mathrm{test}}|}
\sum_{u\in\mathcal{U}_{\mathrm{test}}}
\sum_{i\in\mathcal{I}_u} m_{i,\rho}^{+},\\
\underline{\mathrm{NP}}_{\rho}
&=
\frac{1}{|\mathcal{U}_{\mathrm{test}}|}
\sum_{u\in\mathcal{U}_{\mathrm{test}}}
\sum_{i\in\mathcal{I}_u}
\min\!\left\{0,m_{i,\rho}^{-}-a_{i,\rho}^{+}\right\},\\
\overline{\mathrm{NP}}_{\rho}
&=
\frac{1}{|\mathcal{U}_{\mathrm{test}}|}
\sum_{u\in\mathcal{U}_{\mathrm{test}}}
\sum_{i\in\mathcal{I}_u}m_{i,\rho}^{+}.
\end{aligned}
\label{eq:cross_support_metric_bounds}
\end{equation}
These quantities use the same sum-within-user, average-across-users convention as Eq.~\ref{eq:ces_score}. The revenue/u. upper bound assumes exposure on every transition and uses the largest conditional revenue over the action support. Under $y\leq z$ and $\lambda_{\mathrm{eval}}=1$, net p./u. lies in $[\underline{\mathrm{NP}}_{\rho},\overline{\mathrm{NP}}_{\rho}]$: the lower bound takes the lower conditional-revenue extreme with the largest incentive, or no exposure, whereas the upper allows exposure without payout. These loose bounds are not shaded.

The scorers nearly coincide for BC and IQL ($\lambda=1.5$). For IQL ($\lambda=8.0$), $\mathrm{CES}_{\mathrm{in}}$ overestimates both cost/u. and revenue/u. relative to $\mathrm{CES}_{\mathrm{all}}$; the cost error dominates, so it underestimates net p./u. This diagnostic examines bucket-average actions rather than state-wise OOD for every $(s,a)$ pair. Together with Section~\ref{sec:ces_reliability}, it supports local CES ranking while controlling policy OOD through hyperparameters such as $\lambda$.

\subsection{WM Reliability: Rollout Diagnostics}
\label{app:wm_rollout_diagnostics}

Section~\ref{sec:wm_reliability} reports the quantitative one-step and autoregressive prediction results. The cumulative-error and visual analyses below complement those results.

\subsubsection{Quantitative WM Rollout Error}

\begin{figure}[h]
  \centering
  \includegraphics[width=\linewidth]{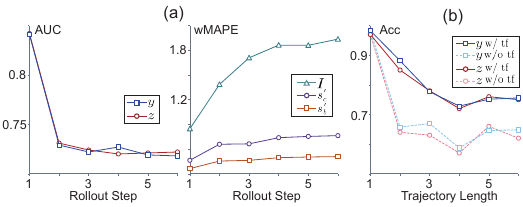}
  \caption{Autoregressive WM rollouts accumulate error, and the degradation is already clear after one to two rollout steps. Panel (a) aggregates prediction performance by rollout step; panel (b) aggregates prediction performance by logged trajectory length. TF denotes teacher forcing.}
  \Description{World-model prediction performance aggregated by rollout step and logged trajectory length.}
  \label{fig:wm_reliability}
\end{figure}

Figure~\ref{fig:wm_reliability} analyzes cumulative WM error from two complementary aggregation perspectives. Figure~\ref{fig:wm_reliability}(a) aggregates predictions by rollout step and shows that recursive prediction error compounds, with substantial deterioration already after one to two steps. Figure~\ref{fig:wm_reliability}(b) aggregates results by logged trajectory length. The WM is most accurate on trajectories of length one, which correspond to initial incentive attempts that generate no exposure. As trajectory length increases, modeling user behavior becomes more difficult, and even one-step teacher-forced prediction degrades. These results motivate the local, one-step rollout design used in the main method.

\subsubsection{Qualitative WM Rollout Visualization}
\label{app:wm_tsne}

\begin{figure}[h]
  \centering
  \includegraphics[width=\linewidth]{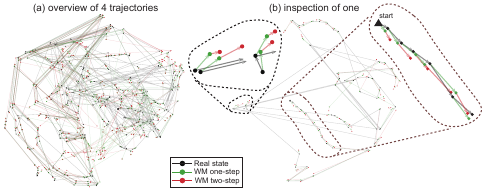}
  \caption{One-step WM predictions stay closer to logged mutable states than selected two-step autoregressive predictions in this qualitative t-SNE view. Points show the mutable subset $s_{\mathrm{rt}}$ of logged states, one-step WM predictions, and two-step autoregressive predictions. t-SNE denotes t-distributed stochastic neighbor embedding.}
  \Description{A t-SNE plot of four trajectories and a zoomed trajectory. Black points are logged mutable states, green points are one-step WM predictions, and red points are two-step autoregressive predictions.}
  \label{fig:wm_tsne}
\end{figure}

Fig.~\ref{fig:wm_tsne} provides a qualitative t-SNE illustration of the same rollout drift quantified in Fig.~\ref{fig:wm_reliability}. It is not used as an OOD test; the quantitative state-error analysis in the main text provides the reliability evidence.

\section{Experimental Setup}
\label{app:exp_config}

The main experimental section evaluates policies on three industrial logged environments and then validates the selected policies online. This appendix records the common experimental substrate: environment provenance and exploration coverage, followed by the shared optimization, rollout, and CES-evaluation protocol. Reliability diagnostics and outcome tables are reported separately in Section~\ref{app:supp_results}.

\subsection{Environment Characteristics}

The industrial offline experiments use three large logged environments that differ in platform source, historical exploration policy, and cost pressure. Table~\ref{tab:environment_config} summarizes their sources and exploration settings. DY-1 and DY-2 use Douyin traffic, whereas DH uses Douyin's Huoshan Version traffic. Although DY-1 and DY-2 share a product source, they come from different serving regimes and budget levels, so they are treated as separate environments. Each environment uses online random exploration around its baseline policy: TD3+BC~\cite{td3bc} for DY-1, DESCN~\cite{UV} for DY-2, and MAB~\cite{mab} for DH. The exploration ranges are relative to the baseline incentive output and are sampled from a uniform random distribution. This randomization is applied at the request level before the incentive value is displayed, matching our request-level optimization target. After the validity filters used for policy training, the three environments contain about 200M, 400M, and 50M transition records, respectively.

\input{tables/environment_config}

All data were preprocessed and filtered according to production validity checks. For both $y$ and $z$, positive samples and negative samples each account for roughly half of the retained records across the datasets.

\subsection{Training Configuration}

\subsubsection{Global Settings}

All offline policy-learning jobs are trained with distributed data parallelism (DDP) on GPUs, each with 32 GB of GPU memory and approximately 16 TFLOPS of FP32 compute. We fix the global batch size to 1024: jobs using 4$\times$ such GPUs use a per-GPU batch size of 256, whereas jobs using 8$\times$ such GPUs use 128. Each candidate policy is trained for 1M optimization steps. Unless stated otherwise below, we use a cosine learning-rate schedule from $10^{-3}$ to $10^{-5}$. The real-time discount in Eq.~\ref{eq:half_life_discount} uses a half-life of 15 minutes.

Each offline configuration is run with three independent random seeds. The user-disjoint train/test split is fixed across seeds. For each seed, we independently train a CES and use it to evaluate the candidate policies trained under the same seed. Reported offline means and standard deviations are the mean and sample standard deviation over these three seed-specific CES evaluation pipelines. Each pipeline aggregates request predictions into per-user metrics according to Eq.~\ref{eq:ces_score}; CES-Fact. applies the same aggregation after evaluating the logged test actions. These settings, including the data split, optimization budget, hardware, and evaluation protocol, are shared across algorithms and are therefore not repeated in the algorithm-specific description below.

\subsubsection{Algorithm-Specific Hyperparameters}

For BC, we minimize the Huber behavior-cloning loss~\cite{huber1964robust}. The 10\% BC baseline is an approximate implementation: in each mini-batch, it applies the same BC objective only to the top 10\% of samples ranked by reward under the $\lambda=1.0$ reward definition. For DT, we follow GAVE~\cite{gave} in treating the decision problem through return-conditioned sequence modeling. In our implementation, the target return-to-go (RTG) is the 60th percentile of realized trajectory returns over user records, and the context length is 6.

For TD3+BC, we set the behavior-cloning coefficient to $\alpha=1.0$. To reduce OOD bias caused by early bootstrap targets, we use a 20K-step warm-up inspired by the official implementation: during the first 10K steps, the actor is trained by pure BC and the Q-function uses a SARSA target; from step 10K to 20K, $\alpha$ is increased from 0 to 1.0 with a cosine schedule, after which the standard TD3+BC update is used. Actions outside the allowed action range are clipped so that their Q-gradient is not back-propagated. TD3+BC uses the standard twin-Q networks. For IQL, we set the expectile parameter to $\tau=0.7$ and use deterministic IQL+DDPG+BC policy extraction with $\alpha=1.0$. Since its policy-extraction step does not use bootstrapping, IQL uses no additional warm-up; its implementation follows the official IQL repository. IQL uses twin Q-functions, with one value network and one actor.

For CQL, the conservative penalty uses 15 proposal actions per state: five actions sampled uniformly, five actions sampled around $\pi(s)$, and five actions sampled around $\pi(s')$. We follow the official CQL repository for the remaining implementation details. For WM-augmented methods (MB TD3+BC, MB IQL, and MB CQL), the synthetic-transition ratio is zero during the 200K-step real-data warm-up and then increases by a cosine schedule to a maximum of 0.5. We use one-step rollouts, refresh them every 100 optimization steps, and set the conservative-loss temperature to 1.0. MB CQL uses the original COMBO algorithmic objective with our structured WM inputs and rollout rules. The conservative penalty uses the same proposal-action construction as CQL. Since COMBO has no official repository, we follow the official CQL implementation together with community implementations.

During WM rollouts, we first sample the exposure indicator from a Bernoulli distribution parameterized by the predicted exposure probability and, only when exposure is sampled, sample the completion indicator from its predicted conditional probability. We set the synthetic attempt-level revenue $I$ to zero when $z=0$ and to the conditional-revenue prediction $\hat m_{\mu,I}$ when $z=1$, then compute the sampled reward as $r^\lambda=I-\lambda ya$ according to Eq.~\ref{eq:immediate_reward}. The same sampled exposure gate gives the done signal: $z=0$, i.e., no user click or acceptance, terminates the trajectory. This causal ordering preserves the structural constraint $y\leq z$ in mixed data. During CES evaluation, in contrast, we use the probabilities directly to compute each request's expected counterfactual revenue and cost, and then aggregate them by user as in Eq.~\ref{eq:ces_score}; this removes Monte Carlo noise from policy ranking.

CES evaluation uses 8 GPUs, each with 32 GB of GPU memory and approximately 16 TFLOPS of FP32 compute, with a per-GPU inference batch size of 2048 (global batch size 16,384). The held-out Ray dataset is materialized once and reused across candidate policies and checkpoints; this reuse and the inference-only evaluation path are the main computational advantage of attaching CES to a dataset.

\subsubsection{Network Architecture}

All networks use a Transformer feature encoder followed by an MLP prediction head. Tokenized inputs are processed by a position-free Transformer encoder to model cross-feature attention. The encoder uses 3--7 layers of feature processing and comprises four Transformer blocks, each with four attention heads; the resulting representation is mapped to the output by a three-layer MLP. For action-conditioned inputs, the action is first projected by a linear layer to a 32-dimensional embedding aligned with the feature-token embedding space. All policy outputs are mapped to the valid action space with a $\tanh$ function. The world model additionally uses an MMoE module with six experts, each having hidden sizes $[64,64]$.

\subsubsection{Feature Engineering}

We partition the raw features into numerical and categorical features. Numerical features are continuous-valued fields. Categorical features include Boolean and string fields, as well as numerical fields with fewer than five categories. Each numerical feature is accompanied by a corresponding bucketized feature to strengthen categorical representations. Thus, for a record with $a$ numerical features and $b$ categorical features, the processed state has $2a+b$ feature dimensions.

All numerical features are normalized by first clipping values to their 1st--99th percentile range and then applying a sigmoid transformation to map them into $[0,1]$. A bucketized numerical feature has at most 128 categories, with excess bins merged. Actions are transformed with Yeo--Johnson transformation, whereas rewards use min--max normalization.

\section{D4RL Benchmark Results}
\label{app:d4rl}

The preceding appendices focus on the industrial task and its deployment conditions. For comparison on a public benchmark, Table~\ref{tab:d4rl_results} reports D4RL~\cite{d4rl} HalfCheetah normalized returns on the Medium and Medium-Replay dataset types. The public baseline and COMBO entries are taken from Park and Lee~\cite{leq} (arXiv v1, Table 3), which reports scores averaged over 100 evaluation trials and cites the respective papers for prior-work results. The MB IQL and MB TD3+BC entries are based on our implementations and test results.

D4RL and incentivized advertising represent two distinct offline-RL settings. HalfCheetah uses long-horizon control trajectories with directly observed scalar rewards. Our industrial episodes consist of a few incentive attempts with business-driven termination, exposure--completion gates, posterior advertiser revenue, and incentive payouts. Accordingly, the D4RL experiments test the generic model-based extensions under a standard transition-and-reward setup. The industrial experiments evaluate the domain-structured WM, offline request context $\epsilon$, and cost-alignment control. The D4RL results provide supplementary evidence of the extensions' viability on public data.

For this benchmark, we replace the domain-structured WM with a standard MLP-based WM. Apart from this replacement, we reuse the hyperparameters reported above and perform no benchmark-specific tuning. Across the four paired comparisons in Table~\ref{tab:d4rl_results}, the MB extension outperforms its plain counterpart in three cases.

\begin{table}[H]
  \centering
  \caption{D4RL HalfCheetah normalized returns on the Medium and Medium-Replay dataset types. Higher is better.}
  \label{tab:d4rl_results}
  \setlength{\tabcolsep}{4pt}
  \begin{tabular}{lcc}
    \Xhline{0.08em}
    Method & Medium & Medium-Replay \\
    \hline
    BC~\cite{BC} & $43.2$ & $37.6$ \\
    IQL~\cite{iql} & $47.4$ & $44.2$ \\
    TD3+BC~\cite{td3bc} & $48.3$ & $44.6$ \\
    CQL~\cite{cql} & $46.9$ & $45.3$ \\
    MB IQL & $49.1 \pm 1.2$ & $46.2 \pm 1.8$ \\
    MB TD3+BC & $49.9 \pm 0.8$ & $43.1 \pm 1.5$ \\
    COMBO~\cite{combo} & $54.2$ & $55.1$ \\
    \Xhline{0.08em}
  \end{tabular}
\end{table}



\end{document}

%% file: tables/offline_reward.tex
\begin{table*}[t]
  \centering
  \caption{Adding the structured WM consistently improves matched model-free backbones in CES evaluation. CES metrics aggregate request-level predictions per user; MF denotes model-free; WM denotes world model; net p./u. denotes per-user net profit. $\Delta$ is the absolute improvement in percentage points, and Rel. gain is $\Delta$ divided by the MF net p./u.}
  \label{tab:offline_reward}
  \setlength{\tabcolsep}{4pt}
  \renewcommand{\arraystretch}{1.05}
  \begin{tabular}{lllcccc}
    \toprule
    Env. & MF backbone & +WM method (Ours) & MF net p./u. & +WM net p./u. (Ours) & $\Delta$ & Rel. gain \\
    \midrule
    DY-1 & TD3+BC & MB TD3+BC & $+13.8\%$ &  {\textcolor{green!50!black}{$+19.9\%$}} &  {\textcolor{green!50!black}{$+6.1$ pp}} &  {\textcolor{green!50!black}{$+44.2\%$}} \\
       & IQL    & MB IQL    & $+16.1\%$ &  {\textcolor{green!50!black}{$+21.9\%$}} &  {\textcolor{green!50!black}{$+5.8$ pp}} &  {\textcolor{green!50!black}{$+36.0\%$}} \\
       & CQL    & MB CQL    & $+16.8\%$ &  {\textcolor{green!50!black}{$+20.7\%$}} &  {\textcolor{green!50!black}{$+3.9$ pp}} &  {\textcolor{green!50!black}{$+23.2\%$}} \\
    \midrule
    DY-2 & TD3+BC & MB TD3+BC & $+14.1\%$ &  {\textcolor{green!50!black}{$+20.9\%$}} &  {\textcolor{green!50!black}{$+6.8$ pp}} &  {\textcolor{green!50!black}{$+48.2\%$}} \\
       & IQL    & MB IQL    & $+15.9\%$ &  {\textcolor{green!50!black}{$+20.8\%$}} &  {\textcolor{green!50!black}{$+4.9$ pp}} &  {\textcolor{green!50!black}{$+30.8\%$}} \\
       & CQL    & MB CQL    & $+17.0\%$ &  {\textcolor{green!50!black}{$+19.2\%$}} &  {\textcolor{green!50!black}{$+2.2$ pp}} &  {\textcolor{green!50!black}{$+12.9\%$}} \\
    \midrule
    DH   & TD3+BC & MB TD3+BC & $+16.1\%$ &  {\textcolor{green!50!black}{$+23.8\%$}} &  {\textcolor{green!50!black}{$+7.7$ pp}} &  {\textcolor{green!50!black}{$+47.8\%$}} \\
         & IQL    & MB IQL    & $+16.8\%$ &  {\textcolor{green!50!black}{$+23.4\%$}} &  {\textcolor{green!50!black}{$+6.6$ pp}} &  {\textcolor{green!50!black}{$+39.3\%$}} \\
         & CQL    & MB CQL    & $+15.9\%$ &  {\textcolor{green!50!black}{$+24.6\%$}} &  {\textcolor{green!50!black}{$+8.7$ pp}} &  {\textcolor{green!50!black}{$+54.7\%$}} \\
    \midrule
    Avg. & TD3+BC & MB TD3+BC & $+14.7\%$ &  {\textcolor{green!50!black}{$+21.5\%$}} &  {\textcolor{green!50!black}{$+6.9$ pp}} &  {\textcolor{green!50!black}{$+46.8\%$}} \\
       & IQL    & MB IQL    & $+16.3\%$ &  {\textcolor{green!50!black}{$+22.0\%$}} &  {\textcolor{green!50!black}{$+5.8$ pp}} &  {\textcolor{green!50!black}{$+35.5\%$}} \\
       & CQL    & MB CQL    & $+16.6\%$ &  {\textcolor{green!50!black}{$+21.5\%$}} &  {\textcolor{green!50!black}{$+4.9$ pp}} &  {\textcolor{green!50!black}{$+29.8\%$}} \\
    \bottomrule
  \end{tabular}
\end{table*}

%% file: tables/online_ab.tex
\begin{table}[t]
  \caption{Sequential online A/B tests support the offline ranking at matched cost: MB IQL gains where plain IQL does not, and removing WM reverses the gain. MB denotes model-based; cost/u. denotes per-user incentive cost; parentheses show $p$-values, and $\sim$0 means $p<0.0001$.}
  \label{tab:online_ab}
  \centering
  \setlength{\tabcolsep}{1.2pt}
  \renewcommand{\arraystretch}{0.95}
  \begin{tabular}{lcccc}
    \toprule
    Metric & Exp. A & Exp. B & Exp. C & Exp. D \\
    \midrule
    Period & Mar. & Mar.--Apr. & May--Jun. & Jun. \\
    Online $\pi$ & DESCN & TD3+BC & TD3+BC & MB IQL \\
    Candidate $\pi$ & TD3+BC & IQL & MB IQL & IQL \\
    Conclusion & {\textcolor{green!50!black}{+Seq. $\uparrow$}} & no diff. & {\textcolor{green!50!black}{+WM $\uparrow$}} & {\textcolor{red!70!black}{-WM $\downarrow$}} \\
    \midrule
    cost/u. & +1.5\% (0.5) & +1.1\% (1.0) & +1.5\% (1.0) & +0.3\% (1.0) \\
    net p./u. & \textbf{\textcolor{green!50!black}{+7.7\% ($\sim$0)}} & -0.9\% (1.0) & \textbf{\textcolor{green!50!black}{+8.0\% ($\sim$0)}} & \textbf{\textcolor{red!70!black}{-6.6\% ($\sim$0)}} \\
    \bottomrule
  \end{tabular}
\end{table}

%% file: tables/ces_reliability.tex
\begin{table}[!h]
  \caption{CES factual prediction with and without request context, and under reduced-support shift. AUC and ECE denote area under the ROC curve and expected calibration error; lower ECE and weighted mean absolute percentage error (wMAPE) are better.}
  \label{tab:ces_reliability}
  \centering
  \begin{tabular}{llcccccc}
    \toprule
    Train & Eval & $\epsilon$
    & \multicolumn{2}{c}{$z$}
    & \multicolumn{2}{c}{$y\mid z$}
    & $I\mid z$ \\
    \cmidrule(lr){4-5}\cmidrule(lr){6-7}
    & & & AUC & ECE & AUC & ECE & wMAPE \\
    \midrule
    DY-2 & DY-2 & N & 0.85 & 0.04 & 0.86 & 0.04 & 0.83 \\ 
    DY-2 & DY-2 & Y & 0.91 & 0.02 & 0.93 & 0.03 & 0.32 \\ 
    \midrule
    DY-2$_{\mathrm{in}}$ & DY-2$_{\mathrm{in}}$  & Y & 0.92 & 0.02 & 0.93 & 0.02 & 0.30 \\ 
    DY-2$_{\mathrm{in}}$ & DY-2$_{\mathrm{out}}$ & Y & 0.78 & 0.07 & 0.79 & 0.07 & 0.99 \\ 
    DY-2$_{\mathrm{in}}$ & DY-2     & Y & 0.83 & 0.06 & 0.84 & 0.05 & 0.61 \\ 
    \bottomrule
  \end{tabular}
\end{table}

%% file: tables/wm_reliability.tex
\begin{table}[t]
  \centering
  \caption{WM prediction with observed request context, context removal, and autoregressive (a.r.) rollout. $\mathrm{wM}$ denotes wMAPE; $\mathrm{c}$ and $\mathrm{b}$ denote CPM- and behavior-type state-field groups. Completion metrics $y $are $y \mid z$, and $I$ denotes $I \mid z$.}
  \label{tab:wm_reliability}
  \setlength{\tabcolsep}{4pt}
  \begin{tabular}{llccccc}
    \toprule
    Input & Eval & $z$ AUC & $y$ AUC & $s'_{\mathrm{c}} \overline{\mathrm{wM}}$ & $ s'_{\mathrm{b}} \overline{\mathrm{wM}}$ & $I$ wM \\
    \midrule
    $(s,\epsilon)$ & 1-step & 0.87 & 0.89 & 0.33 & 0.40 & 0.39\\
    $s$            & 1-step & 0.74 & 0.74 & 0.56 & 0.56 & 0.84\\
    $s$            & a.r.   & 0.70 & 0.68 & 0.69 & 0.61 & 1.41\\
    \bottomrule
  \end{tabular}
\end{table}

%% file: tables/q_regularization_ablation.tex
\begin{table}[t]
  \centering
  \caption{Conservative Q-regularization ablation on DY-2. The regularization benefits the model-based methods but slightly degrades their model-free backbones. Entries report the mean and standard deviation of normalized CES net p./u. across three random-seed runs.}
  \label{tab:q_regularization_ablation}
  \setlength{\tabcolsep}{3pt}
  \renewcommand{\arraystretch}{1.05}
  \begin{tabular}{lcccc}
    \toprule
    & TD3+BC & MB TD3+BC & IQL & MB IQL \\
    \midrule
    \makecell[tl]{w/ Q reg.} &
      \makecell[t]{$+12.6\%$\\$\pm 0.7\%$} &
      \makecell[t]{$+20.9\%$\\$\pm 1.1\%$} &
      \makecell[t]{$+14.0\%$\\$\pm 1.1\%$} &
      \makecell[t]{$+20.8\%$\\$\pm 1.2\%$} \\
    \makecell[tl]{w/o Q reg.} &
      \makecell[t]{$+14.1\%$\\$\pm 0.7\%$} &
      \makecell[t]{$-2.3\%$\\$\pm 0.9\%$} &
      \makecell[t]{$+15.9\%$\\$\pm 0.7\%$} &
      \makecell[t]{$+3.5\%$\\$\pm 1.0\%$} \\
    \bottomrule
  \end{tabular}
\end{table}

%% file: tables/ope_ab_backtest.tex
\begin{table}[t]
  \centering
  \caption{Offline backtests broadly follow the online deployment signs on the same policy pairs. Each env. is the corresponding pre-launch DY- slice; subscripts denote original weights of the experiment where the model first appeared; DESCN$'$ denotes distillation due to the missing original weight file. CES results aggregate request-level predictions per test user; CES and FQE results are normalized by the online policy.}
  \label{tab:ope_ab_backtest}
  \centering
  \setlength{\tabcolsep}{1.2pt}
  \renewcommand{\arraystretch}{0.95}
  \begin{tabular}{lcccc}
    \toprule
    Metric & Exp. A & Exp. B & Exp. C & Exp. D \\
    \midrule
    Online $\pi$ & DESCN$'_A$ & TD3+BC$_A$ & TD3+BC$_A$ & MB IQL$_C$ \\
    Candidate $\pi$ & TD3+BC$_A$ & IQL$_B$ & MB IQL$_C$ & IQL$_D$ \\
    Env. & DY$_A$ & DY$_B$ & DY$_C$ & DY$_D$ \\ 
    \midrule
    A/B test: cost/u. & $+1.5\%$ & $+1.1\%$ & $+1.5\%$ & $+0.3\%$ \\
    A/B test: net p./u. & $+7.7\%$ & $-0.9\%$ & $+8.0\%$ & $-6.6\%$ \\
    \midrule
    CES: cost/u. & +2.3\% & +2.2\% & +0.9\% & -0.5\% \\
    CES: net p./u. & +9.0\% & -2.4\% & +8.2\% & -8.6\% \\
    \midrule
    FQE: cost/u. & +2.0\% & -0.5\% & +0.2\% & +1.2\% \\
    FQE: net p./u. & +9.5\% & -1.3\% & +3.6\% & -8.0\% \\
    \bottomrule
  \end{tabular}
\end{table}

%% file: tables/ope_common_reference.tex
\begin{table}[t]
  \centering
  \caption{Evaluating deployed checkpoints on the same DY-2 users provides a common-reference offline comparison. CES aggregates request-level predictions per test user. Results are normalized by ground truth on DY-2.}
  \label{tab:ope_common_reference}
  \setlength{\tabcolsep}{2.1pt}
  \renewcommand{\arraystretch}{1.0}
  \begin{tabular}{lccccc}
    \toprule
    Metric & DESCN$'_A$ & TD3+BC$_A$ & IQL$_B$ & MB IQL$_C$ & IQL$_D$\\
    \midrule
    CES: net p./u. & -2.0\% & +1.8\% & +2.5\% & +7.9\% & +2.1\%\\
    CES: cost/u.   & -3.7\% & +2.4\% & -0.2\% & +1.5\% & -0.7\%\\
    \midrule
    FQE: net p./u. & -1.5\% & +3.0\% & +3.1\% & +8.4\% & +8.0\%\\
    FQE: cost/u.   & -0.6\% & +2.1\% & +1.0\% & -0.3\% & +0.5\%\\
    \bottomrule
  \end{tabular}
\end{table}

%% file: tables/one_step_ces.tex
\begin{table}[t]
  \caption{Sequential policies score higher than their one-step counterparts on DY-2 in this reference evaluation. CES metrics aggregate request-level predictions per test user.}
  \label{tab:one_step_ces}
  \centering
  \begin{tabular}{lcccc}
    \toprule
    Metric & \shortstack{TD3+BC\\$\gamma=0$} & \shortstack{TD3+BC\\half-life $\gamma$} & \shortstack{IQL\\$\gamma=0$} & \shortstack{IQL\\half-life $\gamma$} \\
    \midrule
    CES: net p./u. & +9.4\% & +11.2\% &  +12.6\%  & +15.0\% \\
    CES: cost/u.   & +1.5\% & +2.9\%  &  +1.0\%  & +0.3\%\\
    \midrule
    FQE: net p./u. & +6.3\% & +9.8\%  &  +8.5\%  & +12.6\%    \\
    FQE: cost/u.   & -1.1\% & -0.1\%  &  +0.9\%  & +2.7\%    \\
    \bottomrule
  \end{tabular}
\end{table}

%% file: tables/one_step_ab.tex
\begin{table}[t]
  \caption{TD3+BC improves net profit over one-step causal or bandit online policies in these reference A/B tests. Revenue/u. denotes per-user revenue; Douyin-HS abbreviates Douyin's Huoshan Version.}
  \label{tab:one_step_ab}
  \centering
  \setlength{\tabcolsep}{1.2pt}
  \renewcommand{\arraystretch}{0.95}
  \begin{tabular}{lcc}
    \toprule
    Metric & Exp. A & Exp. F \\
    \midrule
    Period & Mar. & May \\
    Platform & Douyin & Douyin-HS \\
    Online $\pi$ & DESCN & MAB \\
    Candidate $\pi$ & TD3+BC & TD3+BC \\
    \midrule
    net p./u. ($p$) & \textbf{\textcolor{green!50!black}{+7.72\% ($\sim$0)}} & \textbf{\textcolor{green!50!black}{+10.23\% (0.01)}} \\
    cost/u. ($p$) & +1.54\% (0.45) & +1.82\% (0.78) \\
    revenue/u. ($p$) & \textbf{\textcolor{green!50!black}{+6.55\% ($\sim$0)}} & \textbf{\textcolor{green!50!black}{+7.73\% (0.03)}} \\
    Incentive ROI ($p$) & \textbf{\textcolor{green!50!black}{+3.88\% (0.03)}} & \textbf{\textcolor{green!50!black}{+6.03\% ($\sim$0)}} \\
    \bottomrule
  \end{tabular}
\end{table}

%% file: tables/evaluation_efficiency.tex
\begin{table}[h]
  \centering
  \caption{Wall-time overhead of checkpoint evaluation under the same fixed allocation of 8 GPUs, each with 32 GB of GPU memory and approximately 16 TFLOPS of FP32 compute. The CES time excludes its one-time 1-hour dataset-level training cold start.}
  \label{tab:evaluation_efficiency}
  \setlength{\tabcolsep}{5pt}
  \renewcommand{\arraystretch}{1.05}
  \begin{tabular}{lcc}
    \toprule
    Setting & Wall time & Overhead \\
    \midrule
    Training only & 20 h & 0\% \\
    Training + CES & 21 h & 5\% \\
    Training + FQE & 30 h & 50\% \\
    \bottomrule
  \end{tabular}
\end{table}

%% file: tables/offline_reward_full.tex
\begin{table*}[t]
  \centering
  \caption{Full CES offline evaluation keeps the complete baseline, cost diagnostics, and seed-level variation in one table. All metrics first sum request-level predictions within each test user and then average across users. CES-Fact. uses the logged test actions; each entry reports the mean relative change and $\pm$ one sample standard deviation across three random-seed runs.}
  \label{tab:offline_reward_full}
  \setlength{\tabcolsep}{1.6pt}
  \renewcommand{\arraystretch}{1.0}
  \begin{tabular}{ll*{10}{c}}
    \toprule
    Env. & Metric
    & CES-Fact. & BC & 10\%BC & DT & TD3+BC & IQL & CQL
    & MB TD3+BC
    & MB IQL
    & MB CQL \\
    \midrule
    DY-1 & net p./u. & $+0.2\%$ & $+2.0\%$ & $+8.5\%$ & $+12.3\%$ & $+13.8\%$ & $+16.1\%$ & $+16.8\%$ & $+19.9\%$ & $+21.9\%$ & $+20.7\%$ \\
         &          & $\pm 0.2\%$ & $\pm 0.4\%$ & $\pm 0.5\%$ & $\pm 0.8\%$ & $\pm 0.6\%$ & $\pm 0.6\%$ & $\pm 0.7\%$ & $\pm 1.0\%$ & $\pm 1.1\%$ & $\pm 1.0\%$ \\
         & cost/u.   & $-0.7\%$ & $+2.2\%$ & $-1.0\%$ & $+0.9\%$ & $+2.0\%$ & $+0.9\%$ & $+2.4\%$ & $-2.5\%$ & $+1.8\%$ & $-1.3\%$ \\
         &          & $\pm 0.2\%$ & $\pm 0.2\%$ & $\pm 0.5\%$ & $\pm 0.4\%$ & $\pm 0.7\%$ & $\pm 0.6\%$ & $\pm 0.3\%$ & $\pm 0.7\%$ & $\pm 1.0\%$ & $\pm 0.7\%$ \\
    \midrule
    DY-2 & net p./u. & $+1.2\%$ & $+3.1\%$ & $+8.4\%$ & $+12.5\%$ & $+14.1\%$ & $+15.9\%$ & $+17.0\%$ & $+20.9\%$ & $+20.8\%$ & $+19.2\%$ \\
         &          & $\pm 0.2\%$ & $\pm 0.5\%$ & $\pm 0.5\%$ & $\pm 0.8\%$ & $\pm 0.7\%$ & $\pm 0.7\%$ & $\pm 0.5\%$ & $\pm 1.1\%$ & $\pm 1.2\%$ & $\pm 1.3\%$ \\
         & cost/u.   & $-0.9\%$ & $+1.2\%$ & $-2.1\%$ & $+1.7\%$ & $+2.6\%$ & $+1.4\%$ & $+1.5\%$ & $-2.9\%$ & $+1.1\%$ & $-1.0\%$ \\
         &          & $\pm 0.1\%$ & $\pm 0.4\%$ & $\pm 0.5\%$ & $\pm 0.8\%$ & $\pm 0.6\%$ & $\pm 0.5\%$ & $\pm 0.7\%$ & $\pm 1.3\%$ & $\pm 1.3\%$ & $\pm 1.0\%$ \\
    \midrule
    DH   & net p./u. & $+0.7\%$ & $+2.2\%$ & $+12.1\%$ & $+19.7\%$ & $+16.1\%$ & $+16.8\%$ & $+15.9\%$ & $+23.8\%$ & $+23.4\%$ & $+24.6\%$ \\
         &          & $\pm 0.5\%$ & $\pm 0.6\%$ & $\pm 1.0\%$ & $\pm 1.2\%$ & $\pm 0.9\%$ & $\pm 0.9\%$ & $\pm 1.0\%$ & $\pm 1.4\%$ & $\pm 1.5\%$ & $\pm 1.1\%$ \\
         & cost/u.   & $-2.2\%$ & $+2.0\%$ & $-1.3\%$ & $-2.0\%$ & $+1.5\%$ & $+2.4\%$ & $+1.7\%$ & $-2.6\%$ & $-1.7\%$ & $-2.9\%$ \\
         &          & $\pm 0.3\%$ & $\pm 0.5\%$ & $\pm 0.7\%$ & $\pm 1.0\%$ & $\pm 0.8\%$ & $\pm 0.9\%$ & $\pm 0.8\%$ & $\pm 1.4\%$ & $\pm 1.4\%$ & $\pm 1.3\%$ \\
    \bottomrule
  \end{tabular}
\end{table*}

%% file: tables/online_ab_expanded.tex
\begin{table*}[t]
  \caption{Expanded online A/B metrics.}
  \label{tab:online_ab_expanded}
  \centering
  \setlength{\tabcolsep}{3pt}
  \begin{tabular}{lcccc}
    \toprule
    Metric & Experiment A & Experiment B & Experiment C & Experiment D \\
    \midrule
    Period & Mar & Mar. --Apr.  & May -- Jun.  & Jun.  \\
    Unique DIDs per group & $>10$M & $>10$M & $>10$M & $>10$M \\
    Online policy & DESCN & TD3+BC in Exp. A & TD3+BC in Exp. A & MB IQL in Exp. C \\
    Candidate policy & TD3+BC & IQL & MB IQL & IQL \\
    \midrule
    Per-user send ($p$) & \textbf{\textcolor{green!50!black}{+1.3102\% (<0.0001)}} & \textbf{\textcolor{red!70!black}{-1.1025\% (0.0007)}} & -0.1442\% (0.9906) & \textbf{\textcolor{green!50!black}{+21.8060\% (<0.0001)}} \\
    Per-user exposure ($p$) & \textbf{\textcolor{green!50!black}{+1.4441\% (<0.0001)}} & \textbf{\textcolor{red!70!black}{-1.2222\% (0.0003)}} & -0.0144\% (0.9999) & \textbf{\textcolor{green!50!black}{+23.6790\% (<0.0001)}} \\
    Per-user cost ($p$) & +1.5439\% (0.4543) & +1.0783\% (0.9736) & +1.4978\% (0.9631) & +0.3064\% (0.9990) \\
    Per-user revenue ($p$) & \textbf{\textcolor{green!50!black}{+6.5523\% (<0.0001)}} & -0.5231\% (0.9903) & \textbf{\textcolor{green!50!black}{+6.8004\% (<0.0001)}} & \textbf{\textcolor{red!70!black}{-4.7719\% (0.0002)}} \\
    Per-user net profit ($p$) & \textbf{\textcolor{green!50!black}{+7.7151\% (<0.0001)}} & -0.9185\% (0.9688) & \textbf{\textcolor{green!50!black}{+7.9640\% (<0.0001)}} & \textbf{\textcolor{red!70!black}{-6.5550\% (<0.0001)}} \\
    Incentive ROI ($p$) & \textbf{\textcolor{green!50!black}{+3.8833\% (0.0327)}} & -0.9446\% (0.9853) & +4.3688\% (0.2620) & \textbf{\textcolor{red!70!black}{-5.3171\% (0.0002)}} \\
    \midrule
    LT 7d ($p$) & +0.0109\% (0.2219) & -0.0189\% (0.5980) & +0.0014\% (0.9999) & +0.0116\% (0.5102) \\
    LT 30d ($p$) & +0.0097\% (0.1474) & -0.0180\% (0.4982) & +0.0042\% (0.9460) & +0.0063\% (0.7716) \\
    Cumulative LTV ($p$) & +0.1502\% (0.0751) & -0.0940\% (0.9694) & +0.1332\% (0.2049) & -0.0277\% (0.9888) \\
    \bottomrule
  \end{tabular}
\end{table*}

%% file: tables/environment_config.tex
\begin{table*}[t]
  \caption{The three industrial environments cover different platforms, behavior policies, and logged exploration settings. Env. denotes environment; Platform denotes the serving app; Explr. base denotes the baseline policy around which action perturbations are applied; Explr. range denotes the logged action-perturbation range; Valid rec. denotes valid records; $\epsilon$ denotes the availability of request context; and Test ratio denotes the held-out test proportion.}
  \label{tab:environment_config}
  \centering
  \begin{tabular}{lcccccc}
    \toprule
    Env. & Platform  & Explr. base & Explr. range & Valid rec. & $\epsilon$ & Test ratio\\
    \midrule
    DY-1 & Douyin & TD3+BC~\cite{td3bc} & $\pm 50\%$ & $\sim$200M & Y & $\sim$10\%\\
    DY-2 & Douyin & DESCN~\cite{UV} & $\pm 50\%$ & $\sim$400M & Y & $\sim$10\%\\
    DH & Douyin's Huoshan Version & MAB~\cite{mab} & $\pm 70\%$ & $\sim$50M & N & $\sim$10\%\\
    \bottomrule
  \end{tabular}
\end{table*}